\documentclass[letterpaper, 10 pt, journal]{IEEEtran}

\makeatletter
\def\endthebibliography{%
  \def\@noitemerr{\@latex@warning{Empty `thebibliography' environment}}%
  \endlist
}
\makeatother
 
\IEEEoverridecommandlockouts
\usepackage[nocompress]{cite}
\usepackage{amsmath,amssymb,amsfonts}
\usepackage{graphicx}
\usepackage{textcomp}
\usepackage{xcolor}
\usepackage{todonotes}
\usepackage{amsmath}
\usepackage{leftidx}
\usepackage{siunitx}
\usepackage{multirow}
\usepackage{footnote}
\usepackage[utf8]{inputenc}
\usepackage{array}
\usepackage{mdwmath}
\usepackage{mdwtab}
\usepackage{eqparbox}
\usepackage{url}
\usepackage{cite}
\usepackage{amsmath,amssymb,amsfonts}
\usepackage{algorithm2e}
\RestyleAlgo{ruled}
\usepackage{textcomp}
\usepackage{lettrine}
\usepackage[font=small]{caption}
\usepackage{subcaption}
\usepackage[export]{adjustbox}
\usepackage{amsmath}
\usepackage{leftidx}
\usepackage{siunitx}
\usepackage{multirow}
\usepackage{balance}
\usepackage[export]{adjustbox}
\usepackage{booktabs}
\usepackage{hyperref}
\usepackage{balance}

\usepackage{makecell}
\usepackage{color, soul}
\usepackage{censor}

\renewcommand{\arraystretch}{1.00}

\newcommand{\boldparagraph}[1]{\vspace{0.3em}\noindent{\bf #1} }

\def\BibTeX{{\rm B\kern-.05em{\sc i\kern-.025em b}\kern-.08em
    T\kern-.1667em\lower.7ex\hbox{E}\kern-.125emX}}

\begin{document}

%\title{\LARGE \bf Spatio-Temporal Situational Graphs for Efficient Handling of Situational Changes}
%\title{\LARGE \bf Lifelong Situational Graphs: Be Aware of Changes in the Situation}
%\title{\LARGE \bf Lifelong Time-dependent 3D Scene Graphs: Be Aware of Changes in the Situation}
%\title{\LARGE \bf Lifelong Scene Change Awareness through Time-Dependent Factors}
%\title{Temporally Aware Metric-Semantic Factor Graph for Efficient Long-term Changes Handling}
%\title{Temporally-Aware Metric-Semantic Factor Graph for Efficient Situational Changes Monitoring in Dynamic Environments}
%\title{Temporally-Aware Metric-Semantic Factor Graph for Efficient Monitoring of Situational Changes in Dynamic Environments}
%\title{\LARGE Situation-based Integration of Dynamics and Motion Models in Scene Graphs}
%\title{\LARGE 3D Scene-Graph SLAM jointly optimizing dynamic and static entities}
% \title{\LARGE \bf Dynamic Entity-Aware 3D Scene Graph SLAM}
%\title{\LARGE \bf Joint Robot and Dynamic Entities Pose Estimation via 3D Scene Graph SLAM}
% \title{\LARGE \bf Multi-Constraint Modeling of Dynamic Entities in 3D Scene Graphs for Robust SLAM}
% \title{\LARGE \bf \hl{change Constraint-Based Modeling of Dynamic Entities in 3D Scene Graphs for Robust SLAM}}
\title{\LARGE \bf Dynamic Entity-Aware 3D Scene Graph SLAM for Robust Localization in Highly Dynamic Environments}
\title{\LARGE \bf A 3D Scene Graph–Based SLAM Framework with Semantic Motion Priors for Dynamic Environments}
\title{\LARGE \bf Dynamic Scene Graph SLAM with Semantic Motion Modeling}
\title{\LARGE \bf Dynamic Entity-Aware 3D Scene Graph SLAM with Semantic Motion Modeling}
\title{\LARGE \bf DYNEMO-SLAM: Dynamic Entity and Motion-Aware 3D Scene Graph SLAM}

% \blackout{
\author{Marco Giberna, Muhammad Shaheer, Miguel Fernandez-Cortizas, Jose Andres Millan-Romera, \\  Jose Luis Sanchez-Lopez, and  Holger Voos
\thanks{Authors are with the Automation and Robotics Research Group, Interdisciplinary Centre for Security, Reliability and Trust (SnT), University of Luxembourg. Holger Voos is also associated with the Faculty of Science, Technology and Medicine, University of Luxembourg, Luxembourg.
\tt{\scriptsize{\{marco.giberna, muhammad.shaheer, miguel.fernandez, jose.millan, joseluis.sanchezlopez, holger.voos\}}@uni.lu}}% 
\thanks{*
This work was funded by the Fonds National de la Recherche of Luxembourg (FNR) under the project DEFENCE22/IS/17800397/INVISIMARK.}%
% \thanks{*
% For the purpose of Open Access, and in fulfillment of the obligations arising from the grant agreement, the authors have applied a Creative Commons Attribution 4.0 International (CC BY 4.0) license to any Author Accepted Manuscript version arising from this submission.}
}
% }

\maketitle
\begin{abstract} \label{abstract}
Robots operating in dynamic environments face significant challenges due to the presence of moving agents and displaced objects.
Traditional SLAM systems typically assume a static world or treat dynamic as outliers, discarding their information to preserve map consistency. 
As a result, they cannot exploit dynamic entities as persistent landmarks, do not model and exploit their motion over time, and therefore quickly degrade in highly cluttered environments with few reliable static features.
This paper presents a novel 3D scene graph-based SLAM framework that addresses the challenge of modeling and estimating the pose of dynamic entities into the SLAM backend.
Our framework incorporates semantic motion priors and dynamic entity-aware constraints to jointly optimize the robot trajectory, dynamic entity poses, and the surrounding environment structure within a unified graph formulation. 
In parallel, a dynamic keyframe selection policy and a semantic loop-closure prefiltering step enable the system to remain robust and effective in highly dynamic environments by continuously adapting to scene changes and filtering inconsistent observations. 
The simulation and real-world experimental results show a $49.97\%$ reduction in ATE compared to the baseline method employed, demonstrating the effectiveness of incorporating dynamic entities and estimating their poses for improved robustness and richer scene representation in complex scenarios while maintaining real-time performance.
\end{abstract}
\vspace{-12pt}
\section{Introduction}
\label{introduction}
Robots operate in dynamic environments with moving agents and occasionally displaced objects, making traditional SLAM approaches fail due to the lack of reliable static landmarks \cite{bavle_slam_2023}.
Most of these methods either assume the world is static or simplify the dynamic SLAM problem by tracking and processing dynamics separately. Consequently, they rely solely on known-to-be-static features in the environment, failing whenever actual static landmarks are scarce. 
%whenever it gets too clutter and there are no or too few reliable static features.
To mitigate this, a body of work has attempted to incorporate moving \cite{bescos_dynaslam_2021}, quasi-static \cite{deeb_piecewise-deterministic_2022}, or occasionally displaced landmarks \cite{walcott-bryant_dynamic_2012}.
% However, these approaches exploit such cues only over short horizons and do not use potentially movable landmarks to reduce drifts during later revisits.
However, these approaches use such cues only locally and do not maintain movable entities as persistent landmarks that could support loop closures or drift reduction over revisits.
%precisely the capability needed to operate reliably in mixed, long-term dynamic environments.
They are also restricted to a specific motion class and fail to generalize when the environment exhibits mixed dynamics, such as both moving agents and displaced objects. 
A central limitation is the absence of semantic reasoning and prior knowledge to select an appropriate motion model for each detected entity.
% Many works focused on exploiting moving \cite{morris_importance_2024, bescos_dynaslam_2021}, quasi-static \cite{deeb_piecewise-deterministic_2022} or occasionally displaced landmarks \cite{walcott-bryant_dynamic_2012}, although they are limited in assuming constant velocity motion models, they lack in exploiting information coming from past poses, and they are limited to a specific type of motion, incapable of generalize to all of them.
% In this direction, Khronos\cite{schmid_khronos_2024} presents a unified approach to handle both agent dynamics and scene changes resulting from moved objects. 
% However, it does not fully integrate the dynamic entities in the SLAM pipeline to improve self-localization, and it relies solely on raw observation, lacking the integration of any known semantic class-dependent motion models that characterize the detected dynamic entities. 
Additionally, they miss in exploiting any high-level semantic relation between dynamic entities and the environment, such as a desk vertically constrained to the unmovable floor, which could further improve the map reconstruction accuracy.
This suggests there is a need for models that capture both motion and high-level semantics to inform the SLAM back-end.

\begin{figure}[!t]
    \centering
    \includegraphics[width=\columnwidth]{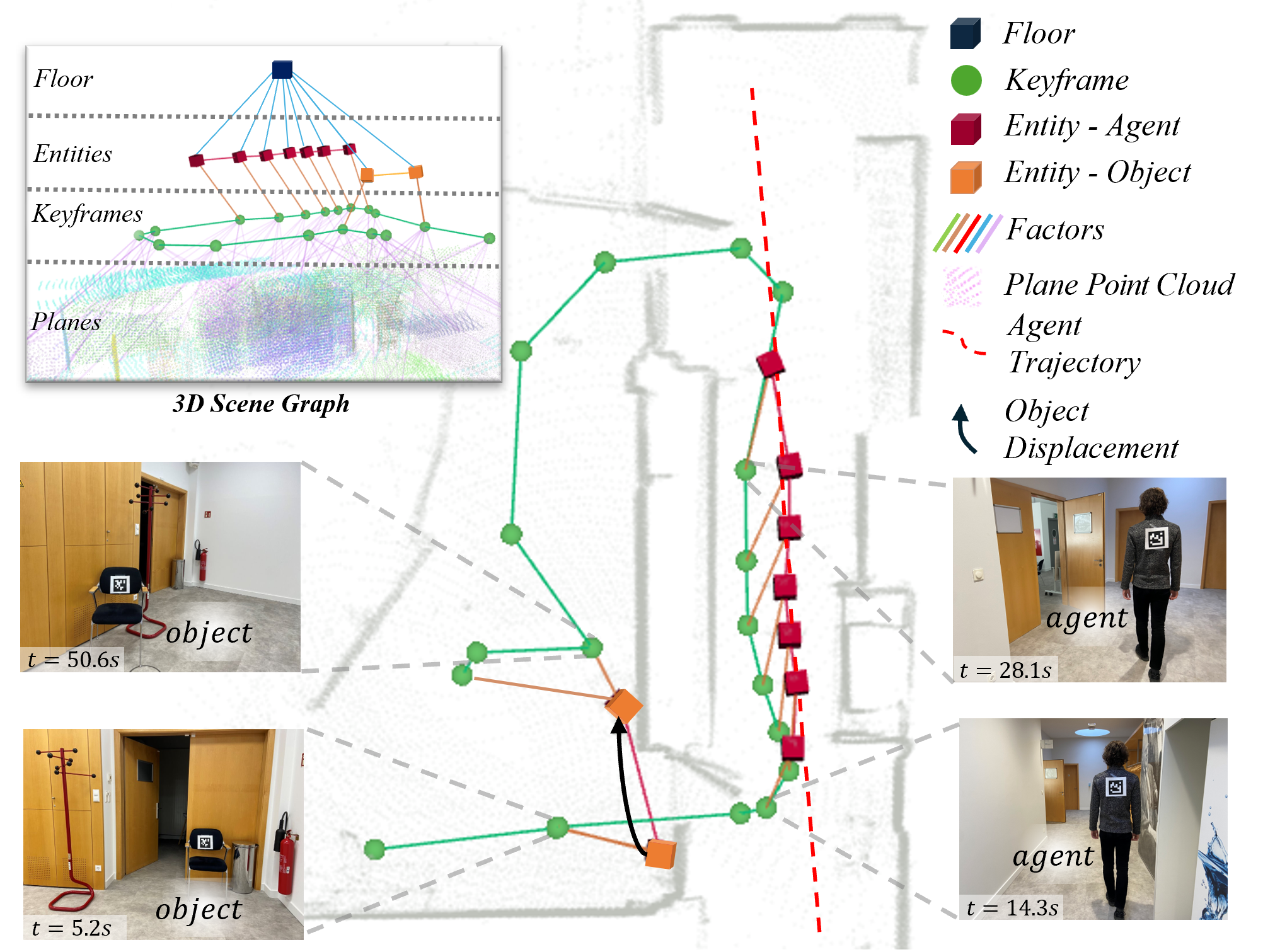}
% \caption{Our approach models dynamic environments by integrating dynamic entities (a walking human and a displaced chair in the figure) into a hierarchical SLAM factor graph. This graph consists of keyframes (green nodes), dynamic entities (red nodes), and floors (blue nodes). Factor graph constraints (colored edges) link entity observations to the keyframes where they were detected from (orange edges), to the eventual previous observation of the same entity (red edges), and to the floor where they were observed (blue edges). Image frames corresponding to selected keyframes are shown. }
% This formulation enables the joint estimation of both the robot’s trajectory and the poses of dynamic entities, leading to a more temporally-consistent and robust scene representation and localization.
\caption{Hierarchical SLAM factor graph integrating dynamic entities (a walking human and a displaced chair) with keyframes, entities, and floor. Image frames show selected keyframes.}
 \label{fig:front_image}
 \vspace{-16pt}
\end{figure}

Recently, SLAM approaches\cite{bavle_s-graphs_2023, hughes_hydra_2022} have leveraged 3D scene graphs to model the environment in terms of different hierarchical levels.
These solutions efficiently categorize, organize, and relate observable instances to higher-level abstractions, like rooms and floors, improving the SLAM accuracy, but assuming the world is static or simplifying the dynamic model by masking or filtering out moving entities. 
%As a consequence, valuable information coming from dynamic entities is neither exploited nor incorporated within the scene graphs. A proper integration of these dynamics is needed to both improve robustness in dynamic environments and enhance the expressiveness of the generated world model.
As a result, valuable information from the dynamic entities remains unused in scene graphs, limiting both robustness in dynamic environments and the expressiveness of the world model.

% To the best of our knowledge, no prior work has leveraged 3D scene graphs tightly coupled with the SLAM backend to jointly model environmental structure, track the robot trajectory, and monitor both dynamic agents and displaced objects while exploiting semantics, high-level relational information, and their features. % to improve loop closure detection.
% % To address these gaps, we present a dynamic graph SLAM framework that incorporates dynamic entities along with their category-wise semantic motion model.
% % Generalizing to both kinds of dynamics that might occur in the scene, these are fully exploited by incorporating entities' observations directly in the world model through mathematical relations, exploiting their semantics to accordingly make use of the most suited motion model. This approach enables joint optimization and consequent improvement of the robot trajectory, entities' pose estimation, and environmental structure simultaneously.
% To address these gaps, we propose a dynamic 3D graph SLAM framework that tightly integrates dynamic entities within the mapping and optimization process.
% Entities are modeled through semantic motion priors and relational constraints, enabling joint optimization of robot trajectory, entity poses, and environmental structure.
To the best of our knowledge, this is the first dynamic 3D graph SLAM (Fig. \ref{fig:front_image}) framework that tightly couples 3D scene graphs with the SLAM backend to jointly model environmental structure, track robot trajectory, and monitor both dynamic agents and displaced objects. This is achieved through semantic motion priors and relational constraints that enable joint optimization of robot trajectory, entity poses, and environmental structure while exploiting semantics, high-level relational information, and their features.
Through experimental validation, we demonstrate that our proposed method is able to generalize from static to highly dynamic environment in a real-time fashion, consistently outperforming the baseline in terms of localization accuracy, while simultaneously estimating poses of the detected dynamic entities. 

% In order to achieve this, we categorize entities based on their motion models, consequentially incorporating it into the scene graphs as factors.
% To empower the entities' motion modeling, and to optimize the quantity and quality of data to be stored, the framework actively detects and registers new information when the \textit{situation changes}, that is, not only accordingly to changes in the state of the robot but also of any the other entities in the scene. % , to overcome loss of information when the explorer is stationary but entities around it are moving.
% Combining semantic-class dependent motion models, high-level semantic-relational constraints, and sparse but rich observations, we improve the pose estimate per instance along with increasing the accuracy of the world model. %, building the foundation to a generalizable world model well suited to predict future entities' states via the embedded history of pose estimates, motion models and high-level semantic relationships.

% List the contributions:
The primary contributions of our paper are as follows.

\begin{itemize}
    \item A \textit{real-time dynamic entity-aware SLAM system} that jointly estimates robot trajectory, static environment structure, and dynamic entity poses, detecting motion, and reusing entities as landmarks.
    \item A \textit{dynamic keyframe selection policy} that registers new keyframes based on significant situational changes, such as robot motion, detection of new entities or pose changes of mapped entities, and incorporates a timer to periodically refresh static observations.
    \item A \textit{set of factors} to constrain dynamic entities with robot keyframes; multiple observations of the same entity to estimate its pose over time via semantic-dependent motion models; and dynamic entities with the structural environment exploiting hierarchy and high-level abstractions.
    \item A \textit{dynamic situationally-aware enhanced loop closure detection}, which filters loop closure candidates via semantic and temporal consistency, removing only features from moved entities to improve closure precision.  
    % \item The integration of semantically dependent motion models of dynamic entities directly into the 3D scene graph, updated by sparse observations registered according to situational changes. 

    % \item A novel Keyframe registration policy triggered by \texit{situational changes} to ensure high computational efficiency and long-term autonomy. %This policy optimizes sparse observations by incorporating low uncertainty achieved through graph-based optimization.
    % \item Integration of the proposed method within the \textit{S-Graphs+} framework \cite{bavle_s-graphs_2023}, along with validation in simulated and real datasets with relevant ablations.
\end{itemize}

\vspace{-12pt}
\section{Related work}
\label{related_work}
% two subsection: dynamic SLAM and scene graphs in dynamic environments
\subsection{Dynamic SLAM}
Research in dynamic SLAM has advanced significantly, with most works \cite{bescos2018dynaslam, song2022dynavins, pfreundschuh_dynamic_2021} focusing on detecting and filtering out dynamic entities to preserve the consistency of static-world SLAM systems, thereby losing valuable data and failing in highly-dynamic environments. 
Beyond filtering, some methods attempt to explicitly model and track dynamic elements. Object-level pose tracking approaches \cite{long2021rigidfusion, strecke2019fusion, behrens_lost_2024} have been proposed, attempting to unify SLAM and tracking process in a coupled framework. However, they miss in reasoning about the semantics of the tracked entities and using prior knowledge about their possible motion capabilities for improving tracking and consequentially the SLAM process.
%although they have been largely limited to a table-top environment with limited success in larger indoor spaces. 
To this end, other approaches \cite{gao_multi-mask_2024, peng_sts-slam_2024} classify landmarks into static, dynamic, or potentially dynamic categories, allowing probabilistic reasoning about their stability but using this information exclusively to mask out unreliable moving features, missing in exploiting their moving features integrated with their inferred motion profiles in subsequent observations and revisits to improve self-localization and mapping.  

A complementary line of research explores the use of moving or displaced entities as landmarks themselves within a graph optimization problem. 
Humans and other agents have been modeled as continuously moving landmarks whose trajectories can be jointly estimated with the robot’s pose \cite{qiu2022airdos, bescos_dynaslam_2021, morris_importance_2024}. 
Similarly, displaced objects have been exploited as scene-consistent features when their motion is sporadic  \cite{ walcott-bryant_dynamic_2012} or quasi-static \cite{deeb_piecewise-deterministic_2022}. 
These works highlight the potential of enriching SLAM beyond static landmarks by leveraging entities that follow different temporal motion profiles. 
However, such existing factor graph formulations typically separate static, dynamic, and potentially dynamic elements, generating independent pose graphs, rather than unifying them into a single representation, hence missing in jointly optimizing entity poses, self-trajectory, and the static background. 
Furthermore, they fail in mixed-dynamics scenarios because of the lack of semantic reasoning to differentiate between possible motions. 
This leaves a gap in exploiting altogether the information provided by both persistently moving agents characterized by known motion models and intermittently displaced objects for long-term mapping and localization.  

Recent advances have begun to address short-term dynamics and long-term environmental changes simultaneously. 
% In this direction, Khronos\cite{schmid_khronos_2024} presents a unified approach to handle both agent dynamics and scene changes resulting from moved objects. 
In this direction, \textit{Khronos} \cite{schmid_khronos_2024} generalizes both the dynamics that occur within and outside the robot’s field of view, presenting a formulation to generate metric-semantic maps that evolve over time. 
However, it does not fully integrate the dynamic entities in the SLAM pipeline to improve self-localization, as it does not model them as factors which relate geometrically and semantically the dynamic object trajectories with those of the robot.
Moreover, it relies solely on raw observation, lacking the integration of any known semantic class-dependent motion models that characterize the detected dynamic entities. 
%Nevertheless, it relies on raw entity observations and does not exploit semantic cues that distinguish between agents and objects, their high-level relations with the environment, or prior knowledge related to their motion models. Furthermore, it misses in leveraging information coming from the observed dynamic entities for improving the localization accuracy.

\vspace{-12pt}
\subsection{3D Scene Graphs for SLAM}
3D scene graphs \cite{armeni_3d_2019, rosinol_3d_2020} efficiently represent environments by encoding semantic abstractions and their relationships in a relational structure. This semantic-rich representation has proven valuable for robust SLAM, enabling geometric and object-level mapping. Recent works such as \cite{kim_3-d_2020, sgf, wald2020learning} use scene graphs for SLAM but only employ flat object-level relationships, ignoring hierarchical structure in higher semantic layers (\textit{rooms, floors}, etc).
Hydra \cite{hughes_hydra_2022} addresses this by generating real-time 3D scene graphs and linking objects with higher-level environmental abstractions such as \textit{rooms}, \textit{floors}, and \textit{buildings}.
They exploit this information to improve loop closure detection and optimize the scene graphs. However, Hydra \cite{hughes_hydra_2022} does not tightly couple the SLAM backend with the generated scene graphs for joint optimization of objects and robot poses. Situational Graphs \cite{bavle_s-graphs_2023} advance the field by implementing a four-layered optimizable factor graph tightly coupling pose estimation and 3D scene graph generation. 
These methods assume that the world is static and generate scene graphs that do not explicitly model the dynamic entities in the environment, furthermore not exploiting any dynamic or potentially dynamic entity to improve loop closure detection. 
\textit{3D DSG} \cite{rosinol_3d_2020} combines semantic reasoning with scene graphs to represent and track dynamic elements, yet not integrating them into the SLAM backend. Moreover, it focuses exclusively on humans, overlooking other moving or repositioned objects.  
%\textit{3D DSG} \cite{rosinol_3d_2020} also represents and tracks dynamic elements but is limited to humans and not integrated into the SLAM backend.
% Hence, these methods cannot exploit the time-dependent states and relations of such entities. % to simultaneously optimize the robot poses and the world model generation. 

% \cite{sgf} achieves real-time segmentation of instances, their semantic attributes, and simultaneous inference of relationships.
% However, their capabilities are limited to inferring geometrical relationships between nearby objects and lack the ability to estimate higher-level concepts like \textit{Floors} and their connections to contained objects.

% \cite{behrens_lost_2024} updates a 3D scene graph and objects-furniture interactions exclusively while actively tracking moving objects within the field of view.

% To the best of our knowledge, no prior work has leveraged 3D scene graphs tightly coupled with the SLAM backend to jointly model environmental structure, track the robot trajectory, and monitor both dynamic agents and displaced objects while exploiting semantics, high-level relational information, and their features to improve loop closure detection.

\begin{figure*}[!htbp]
    \vspace{10pt}
    \centering
    \includegraphics[width=1.0\textwidth]{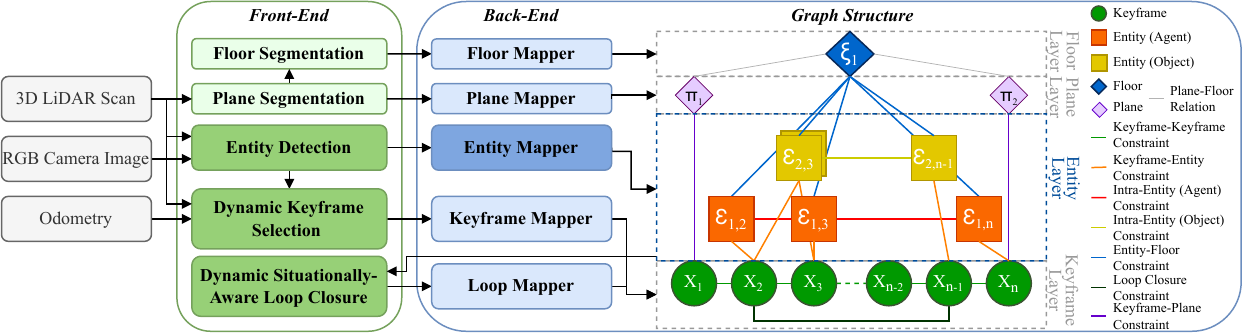}
\caption{\textbf{System Architecture.} The proposed system processes synchronized RGB images, LiDAR scans, and odometry data to build a SLAM graph. The \textbf{front-end} (green) performs entity detection, keyframe selection, plane segmentation, entity-aware loop closure, and floor segmentation. The \textbf{back-end} (blue) registers detected entities and keyframes in the SLAM graph and relative constraints. The \textbf{graph structure} (right) organizes the environment hierarchically, associating keyframes with a detected agent (always observed in new position in the depicted scenario) and object (initially observed twice at the same position, then in a different location), planes and floor. We highlight our contributions (green and blue) with respect to the modules reused from the employed baseline \cite{bavle_s-graphs_2023} (light green, light blue and gray).}
 \label{fig:sys_arc}
\vspace{-18pt}
\end{figure*}
\vspace{-8pt}
\section{Overview}
\label{sec:overview}
\subsection{System Architecture}
% Background
% We exploit the work presented in \cite{bavle_s-graphs_2023} that generates a hierarchical, optimizable scene graph of the environment using a 3D LiDAR sensor.
% Our approach extends this framework by incorporating image data specifically for detecting and modeling dynamic entities in the scene graph. We exploit the floor segmentation module and the keyframe and floor mapper modules from \cite{bavle_s-graphs_2023}. We refine the keyframe selection policy and loop closure detection to account for dynamics, and implement new modules for dynamic entity detection and mapping.

% Overall pipeline
The architecture of the proposed method is illustrated in Fig. \ref{fig:sys_arc}. 
% The system processes synchronized LiDAR scans, image frames, and odometry measurements, which are then fed into the \textit{entity mapper} module. 
% This module extracts dynamic entities and their attributes from raw sensor data, generating a new snapshot for each detected entity instance.
First, the front-end (Sec. \ref{sec:front-end}) of our method processes synchronized LiDAR scans, image frames, and odometry measurements for keyframe selection, plane segmentation and entity detection. The \textit{plane segmentation} module extracts planar surfaces from the LiDAR scans, while the \textit{floor segmentation} module calculates floor center using all the currently extracted planes as in \cite{bavle_s-graphs_2023}. The \textit{entity detection} module (Sec. \ref{sec:entity_detection}) extracts dynamic entities and their attributes, generating a snapshot for each detected entity instance. 
% An entity snapshot at time $t$ contains the entity unique ID $i$, the semantic class $S_i \in \{ agents, objects\} $, estimated entity pose $^{s}{\Tilde{\varepsilon}}_{i,t}$ in the sensor frame $s$ and its related uncertainty $\sigma_{i,t}$, and the point cloud fragment $\psi_{i,t}$.
Based on detected entities and odometry measurements, the \textit{keyframe selection} module (Sec. \ref{sec:keyframe_selection}) decides when a new keyframe should be added to the graph.
Next, the back-end (Sec. \ref{sec:back_end}) registers keyframes, planes, floor and correlated entities in an optimizable hierarchical scene graph comprising keyframe, plane, entity, and floor layers, and jointly optimizes them.
% Next, the back-end (Sec. \ref{sec:back_end}) registers the keyframes and correlated entities in the SLAM graph with appropriate constraints among them and with the floor, segmented and registered by modules taken from \cite{bavle_s-graphs_2023}. 
% Next, based on these keyframes, the entities are added in the SLAM graph and semantic and motion-based constraints are established to corr ctly estimate and optimize their poses.  
% Finally, the graph optimizer refines the reconstructed map jointly with the poses of mapped dynamic entities and robot trajectory.  
Furthermore, the \textit{dynamic entity aware loop closure} module (Sec. \ref{sec:entity_aware_loop_closure}) enhances scan matching–based loop detection by identifying entities with inconsistent poses across candidate keyframes and removing their corresponding point cloud fragments.

 % When a keyframe is registered, the \textit{mapper} module integrates it into the map, introducing new nodes and establishing semantic and motion-based constraints. Specifically, each node is constrained to: its corresponding keyframe; the current floor node; the last observed instance of the same entity,if previously detected.

% An entity database is continuously updated with newly detected entity snapshots. 
% To ensure map consistency, the loop detector module filters out entities that have changed pose from loop closure candidates, thereby preventing conflicts across frames. 
% It then triggers the global optimizer to refine the reconstructed map. 

% The entity database continuously integrates newly detected entity snapshots while maintaining map consistency. The loop detector module systematically filters out entities with changed poses from potential loop closure candidates, preventing cross-frame conflicts. This filtered data is then processed by .

% \textbf{Implementation}
% In order to develop an SLAM framework including the aforementioned properties, 
% The proposed method is built on top of S-Graphs \cite{bavle_s-graphs_2023} and modifies its components to incorporate a new semantic layer that includes dynamic entities, which adds up to the walls and floors layers.  
% For ease of implementation, we exploit fiducial markers attached to the entities we are willing to model, assuming that the association problem across multiple observations is solved.
\vspace{-12pt}
\section{Front-End}
\label{sec:front-end}
\subsection{Entity Detection}
\label{sec:entity_detection}
The entity detection module extracts entities from raw sensor data, classifies them semantically, and associates them with prior observations.
% 
% We use April Tags \cite{wang2016iros} for dynamic object detection and consider the challenges of perception, segmentation, and associating data across multiple observations as already resolved and beyond the scope of this research.
% 
This module estimates the pose of objects along with the related uncertainty, and segments their point cloud representation, obtaining an object fragment, i.e., a partial view of the entity.
Hence, the module outputs an entity snapshot $E_{i,t}$ at time $t$ containing the entity unique ID $i$, the semantic class $S_i \!\!\in\!\! \{ agent, object\} $, estimated pose $^{s}{\Tilde{\varepsilon}}_{i,t}$ in the sensor frame $s$ and its related uncertainty $\sigma_{i,t}$, and the point cloud fragment $\psi_{i,t}$. All detected entities at time $t$ are added to the set  $\mathcal{I}_t$ representing the entity IDs observed at that time.

\vspace{-12pt}
\subsection{Dynamic Keyframe Selection}
\label{sec:keyframe_selection}
% The \textit{keyframe registration policy} utilizes these entity instances alongside odometry measurements to determine whether a new keyframe should be created.
% This module takes the robot odometry and the extracted entities as inputs to create an optimizable factor graph. 
% This graph includes the robot states $^{M}{x}_{R_t}$, $t \in \{1, \dots , T \}$, the entity states $^{M}{\varepsilon}_{E_{i,t}}$, and the environmental model represented as a scene graph. 
% \noindent\textbf{ Registration:}
%The keyframe updater module plays a crucial role in this approach, as new entity observations are added to the map and thus their information stored only when correspondent keyframe are added.
% This module builds upon the keyframe mapper module from \textit{S-Graphs} \cite{bavle_s-graphs_2023}, enhanced with our novel keyframe registration policy.
%Taking in input odometry measurements and, if detected at the current time, the eventual entity snapshot. 
The keyframe selection module determines when a new keyframe, which is a collection of the timestamp $t$, odometry measurement $^{O}{x}_{R_t}$, LiDAR scan $\mathcal{S}_t$ and image frame at time $t$, should be added to the SLAM graph. 
This module is fundamental to our approach, as entity observations are integrated into the map exclusively upon new keyframe registration. 
Our keyframes selection policy is driven by significant \textit{situational changes}, identified as: the robot moving beyond a predefined threshold $\delta_R$ (Eq. \ref{eq:robot_moved}); new entities being detected for the first time (Eq. \ref{eq:new_entity}), and previously mapped entities changing their pose above a threshold $\delta_E$, which depends on the entity pose detection accuracy (Eq. \ref{eq:entity_moved}). 
Hence the situational changes check at current time $T$ is:

\begin{align}
    ||^{O}{x}_{R_{T-1}} \ominus ^{O}{x}_{R_T}|| &> \delta_R \quad \lor\label{eq:robot_moved} \\
    \exists\, i \in \mathcal{I}_T \setminus \mathcal{I}_{\text{map}} &\;\; \quad\qquad \lor \label{eq:new_entity} \\
    ||^{M}{\varepsilon}^{-1}_{i,t} \oplus ^{M}{\Tilde{\varepsilon}}_{i,T}|| &> \delta_E \quad \forall i \in \mathcal{I}_T \label{eq:entity_moved}  
\end{align}
where $^{-1}$ is the inverse operator, $\oplus$ and $\ominus$ are the composition and inverse composition, $||\cdot||$ is the Euclidean norm over the Lia algebra, which accounts for both translational and rotational parts,  ${}^{O}{x}_{R_{T-1}} \in SE(3)$ is the odometry measurement in correspondence of the previous registered keyframe,$\mathcal{I}_{\text{map}}$ is the set of entity IDs already mapped in the SLAM graph, $^{M}{\varepsilon}_{i,t} \in SE(3)$ is the last observation of the same entity available in the graph, and $^{M}{\Tilde{\varepsilon}}_{i,T} = {}^{M}{x}_{O} \oplus {}^{O}{x}_{R_T} \!\oplus\! {}^{R_T}T_{s} \oplus ^{s}{\Tilde{\varepsilon}}_{i,T}$ is the detected entity pose in the map frame. ${}^{M}{x}_{O} \in SE(3)$ models the drift between the odometry frame $O$ and the map frame $M$, and ${}^{R_T}T_{s} \in SE(3)$ is the transformation from the robot's base link to the sensor frame $s$. %, and $^{s}{\Tilde{\varepsilon}}_{i,T} \in SE(3)$ is the estimated entity pose of the new incoming observation in the sensor frame. 

% To mitigate the increasing uncertainty of older entity observations, we introduce a timer-based mechanism as an additional condition. When an entity is re-detected after its timer expires, a new keyframe is registered to capture a fresher and more reliable observation. %, thereby reducing accumulated uncertainty of entities that remain static, ensuring their estimates are periodically updated and remain consistent over time.
% This mechanism not only reduces accumulated uncertainty but also ensures that static entities are periodically re-observed, keeping their estimates up to date and consistent over time.
At this point, entities that remain static are registered again only when they are observed at the same time as any of the aforementioned conditions is triggered.
However, static entities carry valuable information, since repeated observations at the same location can both refine their pose estimate and strengthen the graph via additional closed-loop constraints.
To exploit this, we add a timer-based mechanism as an additional condition: when a static entity is re-detected after its timer expires, a new keyframe is registered. 
This captures a fresher, more reliable observation and reduces the accumulated uncertainty of older detections. %, and ensures that static entities are periodically reinforced within the optimization.

\vspace{-12pt}
\subsection{Dynamic Entity-Aware Loop Closure} \label{sec:entity_aware_loop_closure}
We employ scan matching-based loop closure constraints at keyframe level, following the approach in \cite{bavle_s-graphs_2023}. 
%, propagating the resultant optimization form the robot poses to all the layers of the graph. 
%based on the Normal Distribution Transform-based scan matching. It uses a translational thresholding between the robot pose nodes to identify the loop closure candidates, and optimizes not only the robot poses but all the layers of the S-Graph.
To robustify loop closure detection and to improve scan alignment in the presence of moving agents and objects, we remove exclusively the point cloud fragments of entities that have moved between the candidate keyframes for loop closure.
By querying the graph structure, we identify if the new entity has an inconsistent pose with respect to its previous observations. 
The pose at time $t$ of entity $i$ is considered inconsistent with the new pose at time $T$and therefore added to the set $\textbf{E}_{t,inc}$, if:
\begin{equation}
    ||^{M}{\varepsilon}_{E_{i,t}} \ominus ^{M}{\varepsilon}_{E_{i,T}}|| > \delta'_E \Rightarrow i \in \textbf{E}_{t,inc}    
\end{equation}
where $^{M}{\varepsilon}_{E_{i,t}} \in SE(3)$ is the pose on the map frame of the entity $i$ at time $t$ extracted from the optimized graph, $||\cdot||$ is the Euclidean norm over the Lie algebra, $\delta'_E$ is a distance threshold that accounts for pose estimation accuracy. 

At time $T$, for all pairs of loop closure original candidate scans $\left( \mathcal{S}_t, \mathcal{S}_T\right)$, if the entity $i$, registered to both keyframes, has an inconsistent pose, its previously stored fragment $\psi_{i,t}$ and the new fragment $\psi_{i,T}$ are subtracted from the original scans of the loop closure candidates $\mathcal{S}_t$ and $\mathcal{S}_T$. 
%On the other hand, if an entity has not moved, its previous observations at the same position are maintained, increasing the amount of available features for loop closure detection. 
Hence, the preprocessed scan candidates $\mathcal{S}^{'}_t$ and $\mathcal{S}^{'}_T$ are:
\begin{equation}\label{eq:pc_removal}
\begin{split}
\mathcal{S}^{'}_t &= \mathcal{S}_t \setminus \bigcup_{i \in \textbf{E}_{t,inc}} \psi_{i,t} \quad \forall t=1,\dots,T-1 \\
\mathcal{S}^{'}_T &= \mathcal{S}_T \setminus \bigcup_{i \in \bigcup_{j=t,\dots T-1} \textbf{E}_{j,inc}} \psi_{i,T}
\end{split}
\end{equation}
%where $\mathcal{S}_t$ is the original scan candidate and $\textbf{E}_{t,inc}$ is the set of inconsistent entity's poses at $t$.

By continuously refining the loop closure candidates based on the entity motion, the system mitigates the incorrect loop closure detection and the imprecise scan matching.
%the system improves the global consistency of the reconstructed map while maintaining accurate and up-to-date scene understanding.

\vspace{-12pt}
\section{Back-End}
\label{sec:back_end}
The back-end is responsible for adding nodes and constraints in the graph, and optimizing the global state. Overall, our global optimization state is defined as:
\begin{multline}
\label{eq:global_state}
    s = \left[ {}^{M}{x}_{R_1}, \dots, {}^{M}{x}_{R_T}, {}^{M}
    {\varepsilon}_{E_{1,1}}, \dots, {}^{M}{\varepsilon}_{E_{N,T}}, \right. \\
    {}^{M}{\pi}_{1}, \dots, {}^{M}{\pi}_{P},
    \left. {}^{M}{\xi}_{1}, \dots, {}^{M}{\xi}_{F}, {}^{M}{x}_{O} \right] ^{\top}
\end{multline}
where $^{M}{x}_{R_t} \!\!\in\!\! SE(3)$, $t \!\!\in\!\! \{ 1, ..., T \} $ are the robot poses at $T$ selected keyframes, $^{M}{\varepsilon}_{E_{i,t}} \!\in\! SE(3)$, $i \!\in\! \{1, ... , N\}$, $t \!\in\! \{ 1, ..., T \} $ are the poses of the $i^{th}$ entity at its observed keyframes,${}^{M}{\pi}_{p}$, $p\! \in \! \{1, ..., P\}$ are the plane parameters of the $P$ planes in the scene as in \cite{bavle_s-graphs_2023}, $^{M}{\xi}_{f} \!\in\! \mathbb{R}$, $f \!\in\! \{ 1, ..., F \}$ are the $F$ floors levels, and $^{M}x_O$ models the drift between the odometry frame $O$ and the map frame $M$.
% defined in Eq. \ref{eq:global_state}. 

\vspace{-12pt}
\subsection{Entity Mapper}\label{sec:mapper}
When a keyframe is registered while one or more entities are in the field of view of the robot, the detected entities are mapped.
Given the detected entity pose $^{s}{\Tilde{\varepsilon}}_{i,t}$ in the sensor frame $s$ and the known transformation from the robot’s base link to the sensor frame $^{R_t}T_{s}$, the pose in the robot frame at time $t$ of the entity $i$ is computed as: $ ^{R_t}\Tilde{\varepsilon}_{E_{i,t}} = {}^{R_t}T_{s} \oplus   {}^{s}{\Tilde{\varepsilon}}_{i,t}$. 
Using the newly registered keyframe $^{M}x_{R_{t}}$, the entity pose is then transformed into the map frame and added to the graph: $^{M}\Tilde{\varepsilon}_{E_{i,t}} = {}^{M}x_{R_{t}} \oplus {}^{R_t}\Tilde{\varepsilon}_{E_{i,t}}$

\textbf{Keyframe-Entity Constraint.}
% The entity observation model is independent on the entity's semantic class. 
Each entity observation is initially constrained to the keyframe from which it was detected.
%The detected pose, transformed in the robot frame, is: $ ^{R_t}\Tilde{\varepsilon}_{E_{i,t}} = {}^{R_t}T_{s} \oplus   {}^{s}{\Tilde{\varepsilon}}_{i,t} $. 
The associated cost function to minimize is:
\begin{multline}
    \label{eq:observation_cost}
    c_{KF-E} \left( {}^{M}x_{R_1}, \dots, {}^{M}x_{R_T}, {}^{M}\varepsilon_{E_{1,1}}, \dots, {}^{M}\varepsilon_{E_{N,T}}  \right) = \\
    \sum\limits_{i=0}^{N} \sum\limits_{t=0}^T|| {}^{M}x^{-1}_{R_t}  \oplus {}^{M}\varepsilon_{E_{i,t}} \ominus {}^{R_t}\Tilde{\varepsilon}_{E_{i,t}} ||^{2}_{\Lambda_{\Tilde{\varepsilon}_{i,t}}}
\end{multline}
where $|| \cdot ||_{\Lambda_{\Tilde{\varepsilon}_{i,t}}} $ is the Mahalanobis distance, and $\Lambda_{\Tilde{\varepsilon}_{i,t}}$ is information matrix associated to the observation of the $i^{th}$ entity at time $t$. The information matrix is derived from the uncertainty extracted by the entity detection module: $\Lambda_{\Tilde{\varepsilon}_{i,t}} = \sigma_{i,t}^{-1}$.

\textbf{Intra-Entity Constraint.}
% The entity motion model depends on the entity's semantic class and prior knowledge of its expected motion. 
The Intra-Entity Constraint constrains two observations of the same entity collected by the robot at different keyframes.
Entity observation nodes are constrained pairwise using an expected motion model $\Tilde{f}_{S_i,P_{E_{i,t}}}$, which is a function of the semantic class $S_i$, and the relative pose between the current and previous entity observations defined as, omitting some indices for clarity, $P_{E_{i,t}} \!\!=\! {}^{M}\varepsilon_{i,t-1} \!\ominus\! ^{M}\Tilde{\varepsilon}_{i,t}$.
The associated cost function is:
\begin{multline}\label{eq:motion_cost}
    c_{E-E} \left( {}^{M}\varepsilon_{E_{1,1}}, \dots, {}^{M}\varepsilon_{E_{N,T}}  \right) = \\
    \sum\limits_{i=0}^{N} \sum\limits_{t=0}^T|| {}^{M}\varepsilon^{-1}_{E_{i,t-1}}  \oplus {}^{M}\varepsilon_{E_{i,t}} \ominus \Tilde{f}_{S_i,P_{E_{i,t}}} ||^{2}_{\Lambda_{\Tilde{f}_{S,E_{i,t}}}}
\end{multline}
where $\Lambda_{\Tilde{f}_{S,E_{i,t}}}$ is the information matrix associated to the motion model $\Tilde{f}_{S,E_{i,t}} \in SE(3)$.

The motion model is defined depending on the extracted entity's semantic class $S_i$ and, when applicable, on the specific entity instance, as agents may exhibit distinct motion behaviors modeled accordingly. 
For objects, the motion model first checks whether the object has moved, accounting for estimation noise. 
If the relative pose remains below a fixed, experimentally determined noise threshold $\nu$, it is assumed static. 
Otherwise, the computed relative pose is employed. 
We hence define the motion model $\Tilde{f}_{obj,E_{i,t}}$ for $i^{th}$ entities in the objects class at time $t$:
\begin{equation}\label{eq:motion_model_objects}
\Tilde{f}_{obj,E_{i,t}} = \left\{
\begin{split}
&I_4    \quad \quad \text{ if } ||{}^M\varepsilon_{E_{i,t-1}} \!\!\ominus\! {}^M\varepsilon_{E_{i,t}}||_{\Lambda_{\Tilde{f}_{S,E_{i,t}}}} \!<\! \nu   \\
&P_{E_{i,t}}  \quad\text{ otherwise } 
\end{split} \right.
\end{equation}
where $||\cdot||_{\Lambda_{\Tilde{f}_{S,E_{i,t}}}}$ is the Mahalanobis distance. 
The motion model for entities in the agent class $\Tilde{f}_{agents,E_{i,t}}$ can be any function which incorporates any prior knowledge of the trajectory of agent $i$ at time $t$.
If this prior is not available, we simply set the motion model:
\begin{equation} 
\Tilde{f}_{agents,E_{i,t}} = P_{E_{i,t}} 
\end{equation}
allowing for tracking the agent's motion evolution over time.
In case of agents following a known straight trajectory since the last observation, the motion model incorporates this prior by projecting the relative pose $P_{E_{i,t}}$ onto the motion subspace defined by the unit direction vector $\mathbf{u}$:
\begin{equation}
    \Tilde{f}_{agents,E_{i,t}} = \left( \mathbf{u}\mathbf{u}^\intercal \right) P_{E_{i,t}} \label{eq:motion_model_straight_trajectory}
\end{equation} 
Only the translational component along $\mathbf{u}$ is retained, while the orthogonal translational and rotational components are set to zero and down-weighted, ensuring the residual reflects motion consistent with the expected linear path.

\textbf{Floor-Entity Constraint.}
%Entities are assumed to limulti-e on the ground, 
It can be assumed that entities within the same floor keep a constant distance with respect to the ground, therefore, their $z$ coordinate remain constant over time. 
We hence constrain the entity observation with the current floor node by maintaining constant relative vertical position ${}^{{}^{M}{\xi}_{j}}\Tilde{z}_{E_{i,t}}$ across observations. 
The associated cost function is:
\begin{multline}\label{eq:floor_cost}
    c_{F-E} \left( {}^{M}{\xi}_{1}, \dots, {}^{M}{\xi}_{F}, {}^{M}\varepsilon_{E_{1,1}}, \dots, {}^{M}\varepsilon_{E_{N,T}}  \right) = \\
    \sum\limits_{i=0}^{N} \sum\limits_{t=0}^T||  {}^{M}\xi_{j}  \ominus \left( {}^{M}\varepsilon_{E_{i,t}} \right)_z \ominus {}^{{}^{M}{\xi}_{j}}\Tilde{z}_{E_{i,t}} ||^{2}_{\Lambda_{F-E}}
\end{multline}
where $\xi_{j}$ is the current floor state, $\left( \cdot \right)_z $ denotes the $z$ coordinate, $\Lambda_{F-E}$ is the entity-floor information matrix.

\section{Experimental Evaluation}
\label{results}
To the best of our knowledge, no existing SLAM method jointly models and optimizes the robot trajectory, environmental structure, and the time-varying poses of both objects and agents, rather focusing purely on, e.g., entity reconstruction \cite{schmid_khronos_2024} or  entity tracking \cite{behrens_lost_2024}. 
Furthermore, our framework combines LiDAR with image-based semantic cues, making direct comparison with purely vision-based \cite{bescos2018dynaslam} dynamic SLAM approaches inherently unfair. 
Since no prior work addresses dynamic environments with a comparable multi-sensor and joint-optimization formulation, we restrict our comparison to the baseline we build upon.

\vspace{-14pt}
\subsection{Assumptions}
For ease of implementation, we exploit AprilTag \cite{wang2016iros} fiducial markers for detection of dynamic entities, their pose estimation and classification. This approach allows us to focus on our core contributions while considering the perception, segmentation and data association problem across multiple observations to be solved and outside the scope of this paper.
The uncertainty of the entity detection $\sigma_{i,t}$ is quantified based on the subpixel corner localization error obtained by marker detection.
In our experiments, agents are moving in an almost straight line fashion. 
Therefore, we incorporate this prior into the related motion model as presented in Eq.\ref{eq:motion_model_straight_trajectory}.

\vspace{-14pt}
\subsection{Experimental Setup}
We evaluated our algorithm in both simulated and real environments. All experiments were performed using a laptop computer with an Intel i9-12900H (8 cores, 2.5 GHz) with 32 GB of RAM. 
In all our experiments, the odometry is estimated through LiDAR odometry algorithm as in \cite{bavle_s-graphs_2023}.
The entity’s point cloud fragments are extracted using KD-tree–based Euclidean clustering.

\begin{table}[tbp]
\vspace{6pt}
    \centering
    \renewcommand{\arraystretch}{0.8}
    \scriptsize 
    \caption{\textbf{Ablation Study.} Enabled modules, introduced in Sections \ref{sec:front-end} and \ref{sec:back_end}, for each setup with respect to the baseline. KF, E, and F stand respectively for keyframe, entity and floor.}
    \begin{tabular}{@{\hspace{2pt}}l@{\hspace{2pt}}||ccc|c@{\hspace{6pt}}c@{\hspace{2pt}}c@{\hspace{2pt}}|cc@{\hspace{1pt}}}
        \toprule
        \textbf{Setup} & \multicolumn{8}{c}{\textbf{Contributed Modules}} \\
        %\cmidrule(lr){2-9} 
        & \multicolumn{3}{c|}{\makecell{\textbf{Constraints} \\ \ref{sec:mapper}}} 
        & \multicolumn{3}{c|}{\makecell{\textbf{Entity PC} \\ \textbf{Removal} \ref{sec:entity_aware_loop_closure}}} 
        & \multicolumn{2}{c}{\makecell{\textbf{Keyframe} \\ \textbf{Selection} \ref{sec:keyframe_selection}}} \\
        
        & \rotatebox{90}{KF-E} & \rotatebox{90}{intra-E} & \rotatebox{90}{F-E} 
        & \rotatebox{90}{No} & \rotatebox{90}{Always} & \rotatebox{90}{\makecell{Condi-\\tional}} 
        & \rotatebox{90}{\makecell{Dynamic \\ Policy}} & \rotatebox{90}{Timer} \\
        \midrule
        \textit{Baseline} \cite{bavle_s-graphs_2023} & & & & \checkmark & & & & \\
        \midrule
        \textit{only KF-E} & $\checkmark$ & & & $\checkmark$ & & & $\checkmark$ & \\
        \textit{intra-E} & $\checkmark$ & $\checkmark$ & & $\checkmark$ & & & $\checkmark$ & \\
        \textit{F-E} & $\checkmark$ & & $\checkmark$ & $\checkmark$ & & & $\checkmark$ & \\
        Full Constraint Set (\textit{FCS}) & $\checkmark$ & $\checkmark$ & $\checkmark$ & $\checkmark$ & & & $\checkmark$ & \\
        \midrule
        \textit{always EPCR} & $\checkmark$ & $\checkmark$ & $\checkmark$ & & $\checkmark$ & & $\checkmark$ & \\
        Motion-Based  (\textit{MB-EPCR}) & $\checkmark$ & $\checkmark$ & $\checkmark$ & & & $\checkmark$ & $\checkmark$ & \\
        \midrule
        \textbf{Full System (\textit{Full})} & $\checkmark$ & $\checkmark$ & $\checkmark$ & & & $\checkmark$ & $\checkmark$ & $\checkmark$ \\
        \bottomrule
    \end{tabular}
    \label{tab:ablation_setups}
    \vspace{-19pt}
\end{table}

\boldparagraph{Metrics.} 
We quantitatively assess performance using Absolute Trajectory Error (ATE) comparing trajectories with ground truth to validate our approach's SLAM capabilities in environments with different degrees of dynamicity. 
Additionally, we demonstrate the effectiveness of our approach in jointly estimating and optimizing dynamic entity poses together with the robot trajectory and static background, measuring the error of the entity pose estimation with respect of the groundtruth over time.
Finally, we report the average optimization time for each dataset and setup, and with respect to the number of nodes in the graph to show our approach's real-time capability.

\boldparagraph{Ablation.} 
We assess the contributions of our developed modules by comparing them against each other. 
Table \ref{tab:ablation_setups} summarizes the employed setups and indicates which modules are enabled in each case.
The first four setups differ in terms of which factors are implemented: \textit{\textit{only KF-E}} incorporates only the keyframe-entity constraints; \textit{\textit{intra-E}} also integrates the intra-entity constraints; while \textit{\textit{F-E}} introduces constraints between entities and the floor semantic entity. 
A subset of possible combinations of factors has been selected to ensure graph connectivity.
Finally, Full Constraint Set (\textit{FCS}) setup, includes all proposed factors.
Setups \textit{FCS}, \textit{always EPCR} (Entity Point Cloud Removal) and Motion-Based (\textit{MB-}) \textit{ECPR} analyze the effect of never, always, or conditionally removing dynamic entity point clouds before loop closure detection. 
% \textit{FCS} retains all entity point clouds regardless of motion, while \textit{always EPCR} removes them unconditionally. 
% The motion-based EPCR setup (\textit{MB-EPCR}) selectively removes entity point clouds for loop closure detection only when movement is detected. 
The final configuration (\textit{\textbf{Full}}) further introduces the entity timer within the dynamic keyframe registration policy, which is implemented in every setup but the baseline.

\begin{table}[tbp]
\vspace{6pt}
\centering    
\renewcommand{\arraystretch}{0.8}
\scriptsize
\caption{\textbf{Datasets.} Summary of the employed datasets. Each dataset name encodes the type of dynamics it contains: stationary (\textit{SA-}) or moving agents (\textit{MA-}), and objects that remain static (\textit{-SO}), are relocated (\textit{-MO}) or only rotated (\textit{-RO}).}
\begin{tabular}{@{\hspace{2pt}}l@{\hspace{4pt}}||@{\hspace{4pt}}c@{\hspace{4pt}}c@{\hspace{4pt}}c@{\hspace{4pt}}|@{\hspace{4pt}}c@{\hspace{4pt}}c@{\hspace{4pt}}c@{\hspace{2pt}}}
\toprule
    \textbf{Dataset} & \textbf{\#agents} & \textbf{\#objects} & \textbf{\#moved obj.} & \textbf{Env.} & \textbf{Area} [\si{m^2}] & \textbf{Length} [\si{s}] \\
\midrule
    \textbf{\textit{S-SASO}} & - & $11$ & - & \textit{sim\_1} & $123$ & $373$\\
    \textbf{\textit{S-SAMO}} & - & $11$ & $7$ & \textit{sim\_1} & $123$ & $644$\\
    \textbf{\textit{S-MASO}} & $129$ & $15$ & - & \textit{sim\_2} & $675$ & $351$\\
    \textbf{\textit{S-MAMO}} & $129$ & $15$ & $9$ & \textit{sim\_2} & $675$ & $324$\\
    \textbf{\textit{S-MASO2}} & $51$ & $6$ & - & \textit{sim\_3} & $225$ &  $111$\\
    \textbf{\textit{S-MASO3}} & $51$ & $6$ & - & \textit{sim\_3} & $225$ &  $292$\\
\midrule
    \textbf{\textit{R-SARO}} & - & $8$ & $5$ & \textit{real\_1} & $90$ & $111$\\
    \textbf{\textit{R-SAMO}} & - & $8$ & $5$ & \textit{real\_1} & $90$ & $173$\\
    \textbf{\textit{R-MASO}} & $3$ & $8$ & - & \textit{real\_1} & $90$ & $93$\\
    \textbf{\textit{R-MAMO}} & $3$ & $8$ & $5$ & \textit{real\_1} & $90$ & $324$\\
\bottomrule
\end{tabular}
\label{tab:datasets_summary}
\vspace{-18pt}
\end{table}

\begin{table*}[!htbp]
\vspace{6pt}
    \centering
    \renewcommand{\arraystretch}{1.0}
    \scriptsize 
    \caption{\textbf{Robot's Average Trajectory Error (ATE) Comparison and Computation Time.} Results across six simulated datasets. For each dataset we report RMSE [\si{cm}], Std [\si{cm}], number of detected loop closures (\#LC), and average computation time [\si{ms}]. Lower values are better, except for the number of detected loop closures (\#LC). \textbf{Bold} values are best, and \underline{underlined} values are second-best.}
    \label{tab:sim_experiment_results}
    \resizebox{\textwidth}{!}{
    \begin{tabular}{l||rr@{\hspace{4pt}}r@{\hspace{3pt}}r|rr@{\hspace{4pt}}r@{\hspace{3pt}}r|rr@{\hspace{4pt}}r@{\hspace{3pt}}r|rr@{\hspace{4pt}}r@{\hspace{3pt}}r|rr@{\hspace{4pt}}r@{\hspace{3pt}}r|rr@{\hspace{4pt}}r@{\hspace{3pt}}r}
        \toprule
        \textbf{Setup} 
        & \multicolumn{4}{c|}{\textit{\textbf{S-SASO}}}
        & \multicolumn{4}{c|}{\textit{\textbf{S-MASO}}}
        & \multicolumn{4}{c|}{\textit{\textbf{S-SAMO}}}
        & \multicolumn{4}{c|}{\textit{\textbf{S-MAMO}}}
        & \multicolumn{4}{c|}{\textit{\textbf{S-MASO2}}}
        & \multicolumn{4}{c}{\textit{\textbf{S-MASO3}}} \\
        
        \cmidrule(lr){2-25}
        & \shortstack{RMSE \\ \relax [\si{cm}]}
        & \shortstack{Std \\ \relax [\si{cm}]}
        & \shortstack{\#LC \\ \relax \vphantom{[ms]}}
        & \shortstack{Time\\ \relax [\si{ms}]}
        & \shortstack{RMSE \\ \relax [\si{cm}]}
        & \shortstack{Std \\ \relax [\si{cm}]}
        & \shortstack{\#LC \\ \relax \vphantom{[ms]}}
        & \shortstack{Time\\ \relax [\si{ms}]}
        & \shortstack{RMSE \\ \relax [\si{cm}]}
        & \shortstack{Std \\ \relax [\si{cm}]}
        & \shortstack{\#LC \\ \relax \vphantom{[ms]}}
        & \shortstack{Time\\ \relax [\si{ms}]}
        & \shortstack{RMSE \\ \relax [\si{cm}]}
        & \shortstack{Std \\ \relax [\si{cm}]}
        & \shortstack{\#LC \\ \relax \vphantom{[ms]}}
        & \shortstack{Time\\ \relax [\si{ms}]}
        & \shortstack{RMSE \\ \relax [\si{cm}]}
        & \shortstack{Std \\ \relax [\si{cm}]}
        & \shortstack{\#LC \\ \relax \vphantom{[ms]}}
        & \shortstack{Time\\ \relax [\si{ms}]}
        & \shortstack{RMSE \\ \relax [\si{cm}]}
        & \shortstack{Std \\ \relax [\si{cm}]}
        & \shortstack{\#LC \\ \relax \vphantom{[ms]}}
        & \shortstack{Time\\ \relax [\si{ms}]} \\
        
        \midrule
        Baseline \cite{bavle_s-graphs_2023} 
        & 12.567 & 7.116 & 7 & 37
        & 31.066 & 9.721 & 6 & 12
        & 24.156 & 17.025 & 5 & 24
        & 32.192 & 18.842 & 5 & 10
        & 25.990 & 10.188 & 0 & 10
        & 21.550 & 9.982 & 10 & 9 \\
        \midrule
        \textit{only KF-E}  
        & 9.970 & 5.614 & 7 & 77
        & 24.733 & \underline{9.161} & 6 & 26
        & 13.693 & 11.487 & 7 & 78
        & 18.920 & 10.087 & 6 & 31
        & 24.904 & 16.074 & 0 & 15
        & 18.963 & 9.923 & 11 & 36 \\
        \textit{intra-E}  
        & 9.049 & 5.147 & 7 & 77
        & \textbf{23.945} & 9.900 & 6 & 39
        & 13.746 & 10.052 & 7 & 80
        & 17.705 & 10.059 & 6 & 33
        & 21.963 & 10.606 & 0 & 16
        & 18.493 & 9.595 & 11 & 33 \\
        \textit{F-E}  
        & 9.282 & 5.184 & 7 & 77
        & 24.747 & \textbf{9.073} & 4 & 25
        & 13.175 & 10.568 & 7 & 80
        & 19.051 & 10.246 & 5 & 28
        & 21.031 & 9.302 & 0 & 16
        & 18.379 & 9.473 & 11 & 47 \\
        \textit{FCS}  
        & 7.244 & 5.009 & 7 & 51
        & 24.187 & 9.322 & 6 & 27
        & 13.511 & 9.490 & 7 & 71
        & 16.856 & 9.925 & 6 & 29
        & 21.986 & 10.614 & 0 & 13
        & 17.308 & \underline{7.982} & 11 & 28 \\
        \midrule
        \textit{always EPCR}  
        & 17.375 & 8.882 & 7 & 52
        & 27.437 & 9.605 & 6 & 28
        & 16.281 & 12.446 & 7 & 63
        & \underline{15.313} & 9.678 & 5 & 30
        & 22.074 & 10.719 & 0 & 16
        & 17.476 & 8.202 & 11 & 52 \\
        \textit{MB-EPCR} 
        & \underline{7.176} & \textbf{4.395} & 7 & 58
        & \underline{24.097} & 9.297 & 6 & 26
        & \underline{12.757} & \underline{8.794} & 7 & 71
        & 16.157 & \underline{9.428} & 6 & 28
        & \underline{18.776} & \textbf{7.912} & 1 & 16
        & \underline{17.142} & \textbf{7.833} & 11 & 28 \\
        \midrule
        \textbf{\textit{Full}}      
        & \textbf{6.963} & \underline{4.477} & 8 & 51
        & 24.157 & 9.309 & 6 & 31
        & \textbf{9.045} & \textbf{7.197} & 6 & 78
        & \textbf{15.211} & \textbf{9.226} & 5 & 30
        & \textbf{13.810} & \underline{8.127} & 1 & 25
        & \textbf{17.098} & 8.627 & 11 & 52 \\
        \bottomrule
    \end{tabular}}
\vspace{-18pt}
\end{table*}

\boldparagraph{Simulated Dataset.} 
We evaluate our approach in six simulated (\textit{S-}) datasets, spanning three different kinds of indoor environment, each presenting different dynamics that are moving agents and/or relocated objects. 
Table \ref{tab:datasets_summary} summarizes the characteristics of the datasets. 
\textit{sim\_1} consists in a 7 rooms small plan rich in furniture, used to validate our approach in a static environment and with displaced objects. 
\textit{sim\_2} is composed of three large rooms saturated with moving humans, presenting a total of 15 boxes scattered around, to assess our method in an environment poor of static features. 
\textit{sim\_3} is a large single room saturated with moving humans and 6 static boxes randomly placed, to test our algorithm in extremely cluttered scenarios poor of static features. 
In \textit{S-MASO2} the robot explores the entire room just once, while in \textit{S-MASO3} it repeats the same trajectory three times, to test our approach with and without revisits. 
In datasets with objects displacement (\textit{S-SAMO} and \textit{S-MAMO}), around $60\%$ of the objects are randomly relocated after initial environment exploration. In moving agent scenarios, humans continuously traverse the environment following a straight trajectory. 
All datasets were generated using the Gazebo physics simulator.%, which accurately models the robot platform, LiDAR sensor, RGB camera, and a 3D indoor environment. 

\boldparagraph{Real Dataset.} 
We collected data with a handheld device equipped with Ouster OS1-64 3D LiDAR, a Realsense camera D435i and an Intel Nuc10 (i7 CPU) computer. 
We recorded real (\textit{R-}) datasets in a university hall (\textit{real\_1}) featuring seven chairs arranged throughout the space, with two humans and a legged robot moving among them. 
% Each dataset captures different dynamic scenarios, including stationary (\textit{SA-}) or moving agents (\textit{MA-}), as well as objects that remain static (\textit{-SO}), are relocated (\textit{-MO}) or are rotated (\textit{-RO}). 
Eventual rotation or relocation of objects occur only after the initial environment exploration.
Table \ref{tab:datasets_summary} summarizes the employed datasets. 
% We quantitatively assess the trajectory estimation performance using ATE compared against ground truth.
% Additionally, we precisely measured all object positions at the beginning of each experiment and after any subsequent relocation. All measurements were recorded relative to the starting point of the dataset, which serves as the origin in our reconstructed maps. This ground truth enables us to evaluate the accuracy of pose estimation of our joint optimization approach. %, which simultaneously refines robot trajectory, static environment structure, and dynamic entity poses within a unified framework.
% To assess the accuracy of entity pose estimation, the initial positions of the objects were accurately measured, along with their new locations after being moved, with respect to the starting point of the dataset, which coincides with the generated map's origin. 

\begin{table*}[!htbp]
\vspace{6pt}
    \centering
    \renewcommand{\arraystretch}{1.0}
    \scriptsize 
    \caption{\textbf{Robot's Average Trajectory Error (ATE) Comparison and Computation Time.} Results across four real datasets. For each dataset we report RMSE [\si{cm}], Std [\si{cm}], number of loop closures (\#LC), and average computation time [\si{ms}]. Lower values are better,  except for the number of detected loop closures (\#LC). \textbf{Bold} values are best and \underline{underlined} are second best.}
    \label{tab:real_experiment_results}
    \resizebox{0.7\textwidth}{!}{
    \begin{tabular}{l||rr@{\hspace{4pt}}r@{\hspace{3pt}}r|rr@{\hspace{4pt}}r@{\hspace{3pt}}r|rr@{\hspace{4pt}}r@{\hspace{3pt}}r|rr@{\hspace{4pt}}r@{\hspace{3pt}}r}
    %{@{\hspace{2pt}}l@{\hspace{2pt}}||@{\hspace{2pt}}r@{\hspace{2pt}}r@{\hspace{4pt}}r@{\hspace{3pt}}r@{\hspace{2pt}}|@{\hspace{2pt}}r@{\hspace{2pt}}r@{\hspace{4pt}}r@{\hspace{3pt}}r@{\hspace{2pt}}|@{\hspace{2pt}}r@{\hspace{2pt}}r@{\hspace{4pt}}r@{\hspace{3pt}}r@{\hspace{2pt}}|@{\hspace{2pt}}r@{\hspace{2pt}}r@{\hspace{4pt}}r@{\hspace{3pt}}r@{\hspace{2pt}}}
        \toprule
        \textbf{Setup} 
        & \multicolumn{4}{c|}{\textit{\textbf{R-SARO}}} 
        & \multicolumn{4}{c|}{\textit{\textbf{R-MASO}}} 
        & \multicolumn{4}{c|}{\textit{\textbf{R-SAMO}}} 
        & \multicolumn{4}{c}{\textit{\textbf{R-MAMO}}} \\
        
        \cmidrule(lr){2-17}
        & \shortstack[t]{RMSE \\ \relax [\si{cm}]} 
        & \shortstack[t]{Std \\ \relax [\si{cm}]} 
        & \shortstack[t]{\#LC\\ \relax \vphantom{Time [ms]}}
        & \shortstack[t]{Time \\ \relax [\si{ms}]}
        & \shortstack[t]{RMSE \\ \relax [\si{cm}]} 
        & \shortstack[t]{Std \\ \relax [\si{cm}]} 
        & \shortstack[t]{\#LC\\\relax \vphantom{Time [ms]}}
        & \shortstack[t]{Time \\ \relax [\si{ms}]}
        & \shortstack[t]{RMSE \\ \relax [\si{cm}]} 
        & \shortstack[t]{Std \\ \relax [\si{cm}]} 
        & \shortstack[t]{\#LC\\ \relax \vphantom{Time [ms]}}
        & \shortstack[t]{Time \\ \relax [\si{ms}]}
        & \shortstack[t]{RMSE \\ \relax [\si{cm}]} 
        & \shortstack[t]{Std \\ \relax [\si{cm}]} 
        & \shortstack[t]{\#LC\\ \relax \vphantom{Time [ms]}}
        & \shortstack[t]{Time \\ \relax [\si{ms}]} \\
        
        \midrule
        Baseline \cite{bavle_s-graphs_2023}
        & 17.768 & 13.758 & 1 & 16
        & 8.297  & 3.265  & 1 & 18
        & 13.700 & 4.222  & 1 & 13
        & 9.165  & 4.263  & 2 & 19 \\
        \midrule
        \textit{only KF-E}  
        & 7.352  & \textbf{2.203} & 1 & 68
        & 7.042  & 2.321  & 2 & 117
        & 11.409 & 2.467  & 2 & 203
        & 4.943  & \underline{1.884} & 2 & 113 \\
        \textit{intra-E}  
        & \textbf{5.441} & 2.811 & 1 & 139
        & 8.019  & 2.365 & 2 & 81
        & 10.356 & \underline{2.387} & 2 & 121
        & 5.989  & 2.012 & 2 & 118 \\
        \textit{F-E}  
        & 8.442  & 2.797 & 1 & 107
        & 7.384  & 2.338 & 2 & 86
        & 9.435  & 2.849 & 2 & 125
        & 4.684  & 2.102 & 2 & 108 \\
        \textit{FCS}  
        & 5.583  & 2.947 & 1 & 93
        & 6.607  & 1.749 & 2 & 59
        & 5.404  & 2.911 & 2 & 89
        & 5.002  & 1.933 & 2 & 78 \\
        \midrule
        \textit{always EPCR}  
        & 9.164  & 3.575 & 1 & 96
        & 9.209  & 2.598 & 2 & 76
        & 12.820 & 2.699 & 2 & 81
        & 5.919  & 2.162 & 2 & 82 \\
        \textit{MB-EPCR} 
        & 5.557  & \underline{2.723} & 1 & 76
        & \textbf{3.525} & \underline{1.619} & 2 & 67
        & \textbf{4.470} & \textbf{2.359} & 2 & 49
        & \textbf{3.413} & \textbf{1.601} & 2 & 88 \\
        \midrule
        \textbf{\textit{Full}}      
        & \underline{5.454} & 2.951 & 2 & 86
        & \underline{3.795} & \textbf{1.580} & 2 & 94
        & \underline{4.705} & 2.472 & 2 & 101
        & \underline{4.317} & 2.165 & 3 & 91 \\
        \bottomrule
    \end{tabular}}
    \vspace{-18pt}
\end{table*}

\vspace{-8pt}
\subsection{Results and Discussion}
\boldparagraph{Impact of the Constraints.}
% Robot trajectory estimation
Tables~\ref{tab:sim_experiment_results} and~\ref{tab:real_experiment_results} present the Absolute Trajectory Error (ATE) results obtained in the simulated and real-world experiments.
Across all datasets, adding geometric/semantic constraints consistently improves trajectory accuracy over the baseline. 
Among single-factor variants, \textit{intra-E} and \textit{F-E} generally outperform \textit{only KF-E} by on average $4.62\%$ and $4.02\%$ across all datasets respectively, indicating that (i) exploiting multiple, temporally adjacent observations of the same entity stabilizes the estimate over time, and (ii) anchoring entities to structural semantics (e.g., the floor) provides a strong geometric prior—particularly in scenes with moving agents or sparse static features. 
Delving deeper, we observe how \textit{intra-E} factors are particularly beneficial in presence of moving agents (\textit{MA-}), demonstrating that in such challenging environment, where the baseline struggles the most, anchoring to static landmarks helps the system reducing possible drift induced by the moving entities, being this appreciable also in static objects datasets (\textit{-SO}). Similarly, modeling moving entities expected trajectories directly into the factor graph mitigates potential accumulating error deriving from the detection process. In fact, this is evident in the real experiments (\textit{R-MA-}), whereas the trajectory might deviate more easily with respect to the expected one, thus struggling more in these scenario, although suggesting as a better tuning of the covariances of both motion model and entities detection that better accounts the less-predictable nature of human trajectories might be beneficial to the system. 
When integrating the full set of constraints (\textit{FCS}), the system attains ATE improvements in most simulated sets, and achieves even better performances on real data, confirming that jointly enforcing keyframe–entity, intra-entity, and floor–entity constraints yields the most robust solution. 

\begin{figure}[!tbp]
\centering
% \begin{subfigure}{0.49\columnwidth}
% \centering
%     \includegraphics[width=\columnwidth]{Figures/edge3_entity2_error_position_timer.png}
% \caption{Translational Error}\label{fig:edge3_entity2_pos_err_timer}
% \end{subfigure}
% \begin{subfigure}{0.49\columnwidth}
%     \includegraphics[width=\columnwidth]{Figures/edge3_entity2_error_rotation_timer.png}
%     \caption{Rotational Error}\label{fig:edge3_entity2_rot_err_timer}
% \end{subfigure} \\
\begin{subfigure}{\columnwidth}
\centering
    \includegraphics[width=\columnwidth]{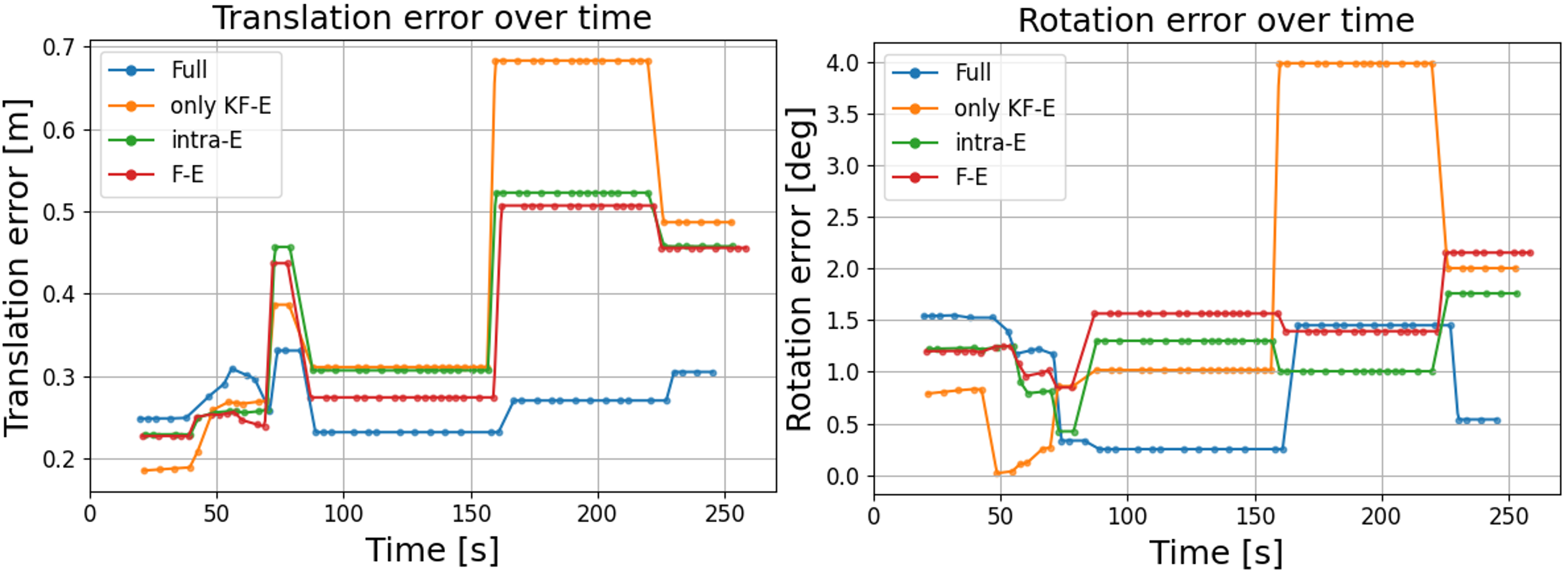}
    \caption{Entity Pose Error - with Timer}\label{fig:edge3_entity10_err_timer}
\end{subfigure} \\
% \begin{subfigure}{0.49\columnwidth}
% \centering
%     \includegraphics[width=\columnwidth]{Figures/edge3_entity13_error_position_timer.png}
% \caption{Translational Error}\label{fig:edge3_entity13_pos_err_timer}
% \end{subfigure}
% \begin{subfigure}{0.49\columnwidth}
%     \includegraphics[width=\columnwidth]{Figures/edge3_entity13_error_rotation_timer.png}
%     \caption{Rotational Error}\label{fig:edge3_entity13_rot_err_timer}
% \end{subfigure} \\
\begin{subfigure}{\columnwidth}
\centering
    \includegraphics[width=\columnwidth]{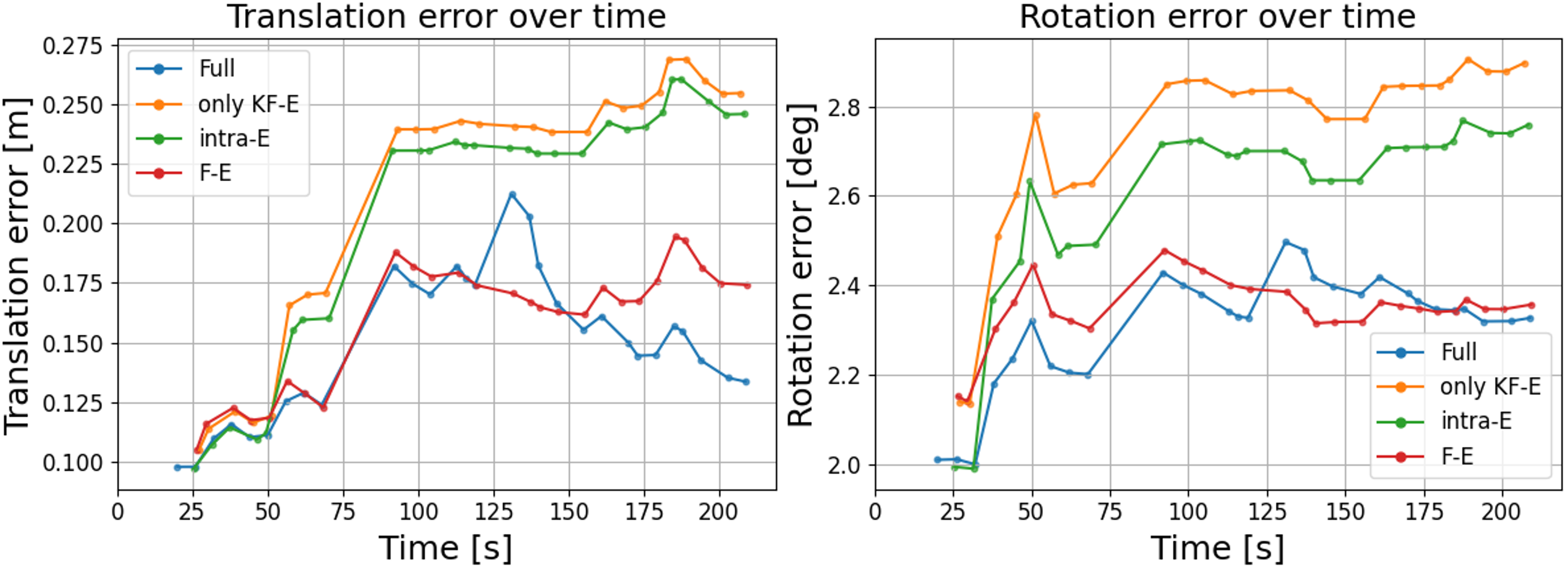}
    \caption{Entity Pose Error - without Timer}\label{fig:edge3_entity13_err}
\end{subfigure}
\caption{\textbf{Entity Pose Estimation Error and Joint Optimization Effect.} Estimated entity pose error, translational and rotational, with respect to the ground truth over time of \textit{\textbf{Full}} method and the setups as presented in Table \ref{tab:ablation_setups} without timer in subfigure \ref{fig:edge3_entity13_err} and all implementing the timer in subfigure \ref{fig:edge3_entity10_err_timer} in the dataset \textit{S-MASO3}.}
\label{fig:entities_pose_estimation_error}
\vspace{-16pt}
\end{figure}

% Entity pose estimation
The entity pose error trends are report in Fig.~\ref{fig:entities_pose_estimation_error}: \textbf{\textit{Full}} starts comparably to other setups but converges to lower translational/rotational errors as more observations are incorporated, while \textit{F-E} is typically second-best due to its semantic anchoring, creating a direct link in the optimization process of the static background with all the entities, and showing how incorporating semantic relations improves the pose estimates.
The \textit{only KF-E} setup performs poorly since the pose estimate relies exclusively on the last observation, while the \textit{intra-E} setup achieves better results over time, especially after new observations come in, weighting all the previous observations.
This shows that keeping multiple recent observations of an entity in the optimization graph improves its pose estimate over time. Although adding new observations to the graph does not necessarily bring an improvement in the pose estimate, we see that an eventual worsening is greatly mitigated by the implementation of intra-entity constraints. Such eventual degradation can be due to less accurate or more noisy entity observations. such as being detected from greater distance or at a steep angle, leading to higher uncertainty.
These results also highlight that keeping multiple recent observations in the graph (\textit{intra-E}) mitigates degradation when new detections are noisier or at challenging viewpoints.

\boldparagraph{Impact of Entity Point-Cloud Removal.}
% Robot trajectory estimation
The effect of pre-processing loop-closure candidates by removing entity point clouds depends on scene dynamics (Tables~\ref{tab:sim_experiment_results} and \ref{tab:real_experiment_results}). 
In static or mostly static environments, \textit{always EPCR} consistently harms performance due to reduced feature availability for place recognition, whereas \textit{FCS} can help by retaining all features.
Conversely, in dynamic scenes (moved objects and/or moving agents), removing entity points before loop closure reduces outlier matches and can bring modest gains. 
Overall, conditionally applying EPCR provides the best trade-off. 
This behavior is reflected by the superior ATE of the \textbf{\textit{Full}} system (\textit{MB-EPCR}), where the dynamic, motion-aware EPCR improves loop-closure quality (accuracy of accepted closures) rather than quantity (\#LC), which often remains similar across EPCR variants. 
This improvement is especially observable in the real datasets, resulting on $24.91\%$ and $54.28\%$ average improvement over \textit{FCS} and \textit{always EPCR}, respectively. In contrast,  performance in the simulated datasets was less pronounced, as the entity point cloud fragment extraction occasionally failed due to the limitations of the implemented segmentation method. Nevertheless, it still achieves $4.93\%$ and $17.12\%$ improvement over \textit{FCS} and \textit{always EPCR}, respectively.

\begin{figure}[!t]
    \centering
    \includegraphics[width=0.95\columnwidth]{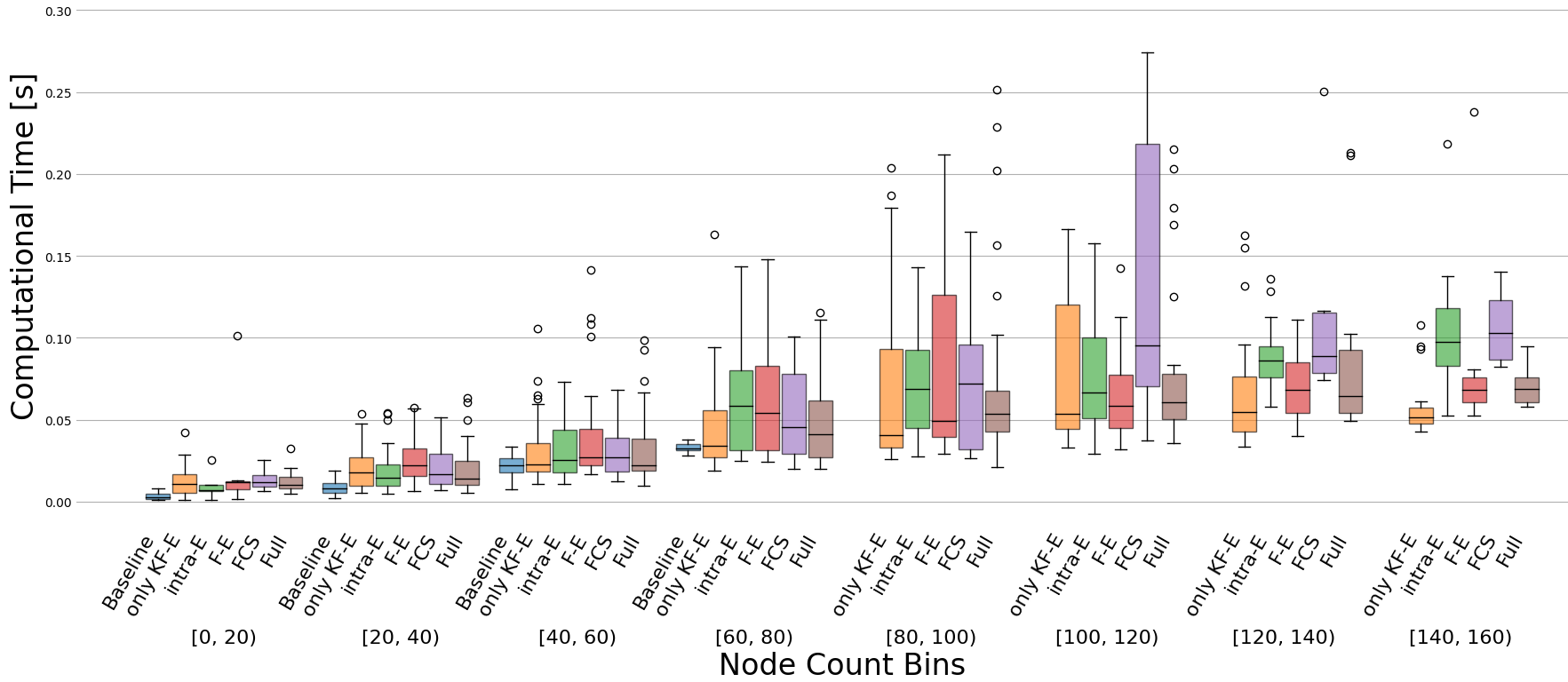}
    \caption{\textbf{Optimization time} [\si{s}] over the number of nodes in the factor graph for each setup and the baseline, whereas data is available, across all employed datasets.}
    \label{fig:computation_time}
\vspace{-22pt}
\end{figure}

\boldparagraph{Impact of the Dynamic Keyframe Updater Policy.}
% Robot trajectory estimation
The dynamic keyframe updater policy increases the likelihood of inserting keyframes when environmental changes occur, improving situation awareness and keeping the graph populated with fresh, informative constraints. 
This is visible from consistent ATE gains of \textit{only KF-E} over Baseline in every dataset, as the main different of this setup with the baseline is the keyframe selection policy, while the additional nodes constrained to the respective keyframe bring minimal new information. 
This is particularly observable in the moved objects (\textit{-MO}) scenarios, as registering new keyframes in correspondence of time instants in which any change in the environment is detected enrich the graph of valuable and up-to-date information through additional keyframes.
A similar trend is observed in the \textit{S-MASO} and \textit{S-MAMO} datasets, where the robot remained stationary at several points. 
Thanks to the dynamic keyframe updater policy, new keyframes were still generated in these positions, not due to robot motion but as a result of changes occurring in the surrounding environment, thus maintaining an up-to-date scene understanding as soon as it restarted.
Our keyframe selection policy tends to increase the quantity of loop-closure detections in dynamic scenarios, as systematically generating  a greater number of keyframes taking in correspondence of situational changes facilitates the loop detection process. 
In simulation, the implementation of the timer mechanism improves the ATE of \textbf{\textit{Full}} over \textit{MB-EPCR} in most datasets and reduced uncertainty through time, improving ATE of $10.22\%$ on average. The timer is particularly beneficial in short trajectories with limited revisits (e.g., \textit{S-MASO2}), where additional entity observations would otherwise be scarce. 
In real data, \textit{MB-EPCR} can match or slightly outperform \textbf{\textit{Full}} in compact environments densely populated with entities, since multiple entities often co-occur in single frames; this naturally yields frequent, fresh observations even without timer-triggered insertions, making the two policies nearly equivalent. 

% Entity pose estimation
Focusing on the entity pose estimation error, Fig.~\ref{fig:entities_pose_estimation_error} shows that when the timer is disabled the graph can be less resilient to transient degradations, whereas the timer helps maintain stability by regularly injecting up-to-date entity constraints.
Such degradations, as well as possible later improvements, are due to the accumulated error in the robot trajectory estimation, which has repercussion on the static environment reconstruction and dynamic entity pose estimation, through the joint optimization procedure.
Table~\ref{tab:real_entity_error} reports the entity pose errors at the time of first detection and after the final optimization in the real-world experiments. 
The results show on average a consistent reduction in error over time, confirming that incorporating additional observations and jointly optimizing robot and entity poses enhances overall estimation accuracy. 
Static objects benefit the most from multiple observations acquired from different points of view, allowing the optimization to leverage the most reliable ones for improved consistency. 
In contrast, the accuracy of moved objects depends on whether a sufficient number of high-quality observations are available before and after motion, highlighting the importance of observation diversity within the graph for achieving robust and accurate estimates.

\begin{table}[!t]
\vspace{6pt}
    \centering
    \renewcommand{\arraystretch}{0.8}
    \scriptsize 
    \caption{\textbf{Entity Pose Estimation Error.} Norm error with respect to ground truth for each entity running \textbf{\textit{Full}} method, when first added to the map and after the last optimization step in the real-world datasets. Entities 4, 5, 6 and 7 are moved in \textit{R-SAMO} and \textit{R-MAMO}, while entities 1, 2 and 3 are rotated in \textit{R-SARO}.}
    \label{tab:real_entity_error}
    \begin{tabular}{@{\hspace{4pt}}c@{\hspace{4pt}}||@{\hspace{4pt}}S@{\hspace{3pt}}S@{\hspace{4pt}}|@{\hspace{4pt}}S@{\hspace{3pt}}S@{\hspace{4pt}}|@{\hspace{4pt}}S@{\hspace{3pt}}S@{\hspace{4pt}}|@{\hspace{4pt}}S@{\hspace{3pt}}S@{\hspace{4pt}}}
        \toprule
        \multirow{2}{*}{\shortstack{\textbf{Entity} \\ \textbf{ID}}} & \multicolumn{2}{c|}{\shortstack{\textbf{\textit{R-SARO}} \\ \relax [\si{cm}]}}& \multicolumn{2}{c|}{\shortstack{\textbf{\textit{R-SAMO}} \\ \relax [\si{cm}]}}& \multicolumn{2}{c|}{\shortstack{\textbf{\textit{R-MASO}} \\ \relax [\si{cm}]}}& \multicolumn{2}{c}{\shortstack{\textbf{\textit{R-MAMO}} \\ \relax [\si{cm}]}} \\ %\cmidrule(lr){2-9}
         & First & Last & First & Last & First & Last & First & Last \\ 
        \midrule
        1 & 16.7 & 9.8 & 20.2 & 6.0 & 7.6 & 6.0 & 10.6 & 8.4 \\
        2 & 2.1 & 5.8 & 5.1 & 5.9 & 2.5 & 2.1 & 10.3 & 4.9 \\
        3 & 1.7 & 8.0 & 6.3 & 2.6 & 7.7 & 5.3 & 6.4 & 5.2 \\
        4 & 13.4 & 11.9 & 6.4 & 2.5 & 2.2 & 3.0 & 9.9 & 8.2 \\
        5 & 8.8 & 4.9 & 8.3 & 5.0 & 5.4 & 0.9 & 5.3 & 2.0 \\
        6 & 9.3 & 6.9 & 2.5 & 6.3 & 8.4 & 1.0 & 0.9 & 4.3 \\
        7 & 4.2 & 3.8 & 2.1 & 1.4 & 3.3 & 0.9 & 7.8 & 4.4 \\
        \midrule
        \textbf{Average} & 8.0 & 7.3 & 7.3 & 4.3 & 4.2 & 1.9 & 7.3 & 5.4 \\
        \bottomrule
    \end{tabular}
    \vspace{-18pt}
\end{table}

\boldparagraph{Computation Time.} 
Tables ~\ref{tab:sim_experiment_results} and~\ref{tab:real_experiment_results} report the computation times for all setups.
The average optimization time in simulated scenarios is $45 \si{ms}$ for \textit{\textbf{Full}} method, and $99\si{ms}$ in real ones, showing that our proposed framework runs real time.

Figure \ref{fig:computation_time} shows the graph optimization time taken by our algorithm compared with the aforementioned setups and baseline with respect to the number of nodes in the graph. 
We can observe as in correspondence of a similar amount of nodes in the graph to be optimized, the type and amount of activated factors speed up the process.
In fact, the optimization time initially increases with respect to the number of nodes consistently across all the approaches, up to showing evident discrepancies starting from over 60 nodes.
The \textit{only KF-E} setup generally achieves the fastest time, because of the absence of multiple constraints (closed loops) on the entity nodes.
Adding floor-entity or, especially, intra-entity constraints increases the computation time as multiple constraints are now effecting the same state, and clearly this recurs in \textit{FCS} setup, whereas both factors occur together. 
Notably, when adding the timer, in \textit{\textbf{Full}} approach, we can generally observe a consistently faster optimization time, suggesting that adding redundant observations and hard constraints on unmoved objects simplifies and speed up the optimization process. 
Moreover, conversely to \textit{FCS} and \textit{intra-E} setups which tend to substantially increase in optimization time for larger graph size, \textit{\textbf{Full}} flattens out, showing even less variability.

\vspace{-10pt}
\section{Conclusion}
We presented a 3D scene graph–based SLAM framework that jointly estimates the poses of both static and dynamic entities, including moved objects and moving agents. 
Comprehensive evaluations show that the best performance of our system is achieved through the combined use of all proposed modules.
The set of factor constraints (\textit{Keyframe–Entity}, \textit{intra–Entity}, and \textit{Floor–Entity}) provides the first substantial improvement in terms of robot trajectory and entity pose estimation by enforcing temporal consistency across observations, exploiting semantic motion prior, and leveraging structural relations with the environment. 
Building on this, the Dynamic Entity-Aware loop-closure filtering introduces an additional gain by removing only those entities that have moved between candidate keyframes, enabling more accurate loop closures in dynamic areas, proving especially effective in highly dynamic and cluttered environment. 
Finally, the dynamic keyframe selection policy with the entity-aware timer brings major improvement by registering keyframes when meaningful situational changes occur, keeping entity observations fresh and limiting temporal uncertainty growth. 
By exploiting sparse yet information-rich cues, the system anchors to the few static landmarks available to mitigate drift, even when the robot remains stationary and the surroundings evolve, achieving robust localization and mapping in challenging scenarios while maintaining real-time performance.
%Comprehensive evaluation show that its best performance is achieved through the combined use of our proposed modules: the factor constraints (\textit{Keyframe–Entity}, \textit{intra–Entity}, and \textit{Floor–Entity}), the Dynamic Entity-Aware loop-closure filtering, and the dynamic keyframe selection policy with the entity-aware timer, which constitute the main contributions of this work. 
% Together, these components improve trajectory accuracy, stabilize entity pose estimates over time, and maintain real-time performance under diverse dynamic conditions. 
% The approach proves particularly effective in highly dynamic and cluttered environments, where reliable static features are scarce. 
% Exploiting sparse observations, it anchors to the few static landmarks available to mitigate drift, even when the robot remains stationary and the surroundings evolve. 
% Moreover, by exploiting known motion and semantic priors, such as linear trajectories and constant height along the $z$-axis through the proposed constraints, point cloud filtering, and keyframe selection strategy, the system achieves robust localization and mapping in challenging scenarios.
% limitations and future works ...
%Current limitations include the reliance on a fixed threshold for dynamic entity motion detection, which hinders the identification of subtle motions, and the use of fiducial markers for perception. 
Future works will focus on enabling marker-free detection and incorporating higher-level semantic layers, e.g., rooms, to enhance scene understanding and to address the entity association problem. In addition, we plan to validate our approach using more complex motion models.

% We presented a 3D scene graph–based SLAM framework that jointly estimates the poses of both static and dynamic entities, including moved objects and moving agents. 
% Building upon ~\cite{bavle_s-graphs_2023}, our approach integrates dynamic entity modeling, motion-aware point cloud filtering, and a dynamic keyframe selection policy. 
% The combined use of multi-factor constraints (\textit{Entity–Keyframe}, \textit{intra–Entity}, and \textit{Entity–Floor}) improves trajectory accuracy, stabilizes and enhances entity pose estimates, and maintains real-time performance across varying levels of dynamicity. 
% Experiments on simulated and real datasets demonstrate superior robustness and accuracy, particularly in highly dynamic or cluttered environments where static features are limited, enabling consistent and temporally aware localization and mapping in complex real-world scenarios.

% \balance
%\clearpage
\vspace{-10pt}
\bibliographystyle{IEEEtran}
\bibliography{bibliography}

@inproceedings{hughes_hydra_2022,
	title = {Hydra: {A} {Real}-time {Spatial} {Perception} {System} for {3D} {Scene} {Graph} {Construction} and {Optimization}},
	isbn = {978-0-9923747-8-5},
	shorttitle = {Hydra},
	doi = {10.15607/RSS.2022.XVIII.050},
	language = {en},
	urldate = {2024-08-01},
	booktitle = {Robotics: {Science} and {Systems} {XVIII}},
	author = {Hughes, Nathan and Chang, Yun and Carlone, Luca},
	year = {2022},
}

@INPROCEEDINGS{pfreundschuh_dynamic_2021,
  author={Pfreundschuh, Patrick and Hendrikx, Hubertus F.C. and Reijgwart, Victor and Dubé, Renaud and Siegwart, Roland and Cramariuc, Andrei},
  booktitle={IEEE International Conference on Robotics and Automation (ICRA)}, 
  title={Dynamic Object Aware LiDAR SLAM based on Automatic Generation of Training Data}, 
  year={2021},
  volume={},
  number={},
  doi={10.1109/ICRA48506.2021.9560730}}

@inproceedings{walcott-bryant_dynamic_2012,
	title = {Dynamic pose graph {SLAM}: {Long}-term mapping in low dynamic environments},
	isbn = {978-1-4673-1736-8 978-1-4673-1737-5 978-1-4673-1735-1},
	shorttitle = {Dynamic pose graph {SLAM}},
	doi = {10.1109/IROS.2012.6385561},
	urldate = {2025-05-19},
	booktitle = {{IEEE}/{RSJ} {International} {Conference} on {Intelligent} {Robots} and {Systems}},
	author = {Walcott-Bryant, Aisha and Kaess, Michael and Johannsson, Hordur and Leonard, John J.},
	year = {2012},
}

@article{deeb_piecewise-deterministic_2022,
	title = {Piecewise-deterministic {Quasi}-static {Pose} {Graph} {SLAM} in {Unstructured} {Dynamic} {Environments}},
	issn = {0921-0296, 1573-0409},
	doi = {10.1007/s10846-022-01739-5},
	language = {en},
	urldate = {2025-05-27},
	journal = {Journal of Intelligent \& Robotic Systems},
	author = {Deeb, Amy and Seto, Mae and Pan, Ya-Jun},
	year = {2022},
}

@inproceedings{morris_importance_2024,
	title = {The {Importance} of {Coordinate} {Frames} in {Dynamic} {SLAM}},
	copyright = {https://doi.org/10.15223/policy-029},
	doi = {10.1109/icra57147.2024.10610840},
	urldate = {2025-07-09},
	booktitle = {{IEEE} {International} {Conference} on {Robotics} and {Automation} ({ICRA})},
	author = {Morris, Jesse and Wang, Yiduo and Ila, Viorela},
	year = {2024},
}

@article{bavle_s-graphs_2023,
	title = {S-{Graphs}+: {Real}-{Time} {Localization} and {Mapping} {Leveraging} {Hierarchical} {Representations}},
	issn = {2377-3766},
	shorttitle = {S-{Graphs}+},
	doi = {10.1109/LRA.2023.3290512},
	journal = {IEEE Robotics and Automation Letters},
	author = {Bavle, Hriday and Sanchez-Lopez, Jose Luis and Shaheer, Muhammad and Civera, Javier and Voos, Holger},
	year = {2023},
}

@inproceedings{armeni_3d_2019,
	title = {{3D} {Scene} {Graph}: {A} {Structure} for {Unified} {Semantics}, {3D} {Space}, and {Camera}},
	shorttitle = {{3D} {Scene} {Graph}},
	doi = {10.1109/ICCV.2019.00576},
	urldate = {2024-08-01},
	booktitle = {{IEEE}/{CVF} {International} {Conference} on {Computer} {Vision} ({ICCV})},
	author = {Armeni, Iro and He, Zhi-Yang and Zamir, Amir and Gwak, Junyoung and Malik, Jitendra and Fischer, Martin and Savarese, Silvio},
	year = {2019},
}

@article{bescos_dynaslam_2021,
  title={DynaSLAM II: Tightly-coupled multi-object tracking and SLAM},
  author={Bescos, Berta and Campos, Carlos and Tard{\'o}s, Juan D and Neira, Jos{\'e}},
  journal={IEEE Robotics and Automation Letters},
  year={2021},
  publisher={IEEE}
}

@article{kim_3-d_2020,
	title = {3-{D} {Scene} {Graph}: {A} {Sparse} and {Semantic} {Representation} of {Physical} {Environments} for {Intelligent} {Agents}},
	issn = {2168-2275},
	shorttitle = {3-{D} {Scene} {Graph}},
	doi = {10.1109/TCYB.2019.2931042},
	urldate = {2024-08-01},
	journal = {IEEE Transactions on Cybernetics},
	author = {Kim, Ue-Hwan and Park, Jin-Man and Song, Taek-jin and Kim, Jong-Hwan},
	year = {2020},
}

@inproceedings{
schmid_khronos_2024,
title = "Khronos: A Unified Approach for Spatio-Temporal Metric-Semantic SLAM
in Dynamic Environments",
author = {Lukas Schmid and Marcus Abate and Yun Chang and Luca Carlone},
year = {2024},
booktitle = "Proc. of Robotics: Science and Systems",
}

@article{bavle_slam_2023,
	title = {From {SLAM} to {Situational} {Awareness}: {Challenges} and {Survey}},
	issn = {1424-8220},
	doi = {10.3390/s23104849},
	journal = {Sensors},
	author = {Bavle, Hriday and Sanchez-Lopez, Jose Luis and Cimarelli, Claudio and Tourani, Ali and Voos, Holger},
	year = {2023},
}

@article{gao_multi-mask_2024,
	title = {Multi-{Mask} {Fusion}-{Based} {RGB}-{D} {SLAM} in {Dynamic} {Environments}},
	copyright = {https://ieeexplore.ieee.org/Xplorehelp/downloads/license-information/IEEE.html},
	issn = {1530-437X, 1558-1748, 2379-9153},
	doi = {10.1109/JSEN.2024.3424469},
	urldate = {2024-07-31},
	journal = {IEEE Sensors Journal},
	author = {Gao, Ye and Hu, Mingnan and Chen, Bo and Yang, Wei and Wang, Jianbin and Wang, Jianzheng},
	year = {2024},
}

@article{peng_sts-slam_2024,
	title = {{STS}-{SLAM}: {Joint} {Visual} {SLAM} and {Multi}-{Object} {Tracking} {Based} on {Spatio}-{Temporal} {Similarity}},
	copyright = {https://ieeexplore.ieee.org/Xplorehelp/downloads/license-information/IEEE.html},
	issn = {2379-8904, 2379-8858},
	shorttitle = {{STS}-{SLAM}},
	doi = {10.1109/TIV.2024.3415006},
	urldate = {2024-10-01},
	journal = {IEEE Transactions on Intelligent Vehicles},
	author = {Peng, Song and Ran, Teng and Zhang, Jianbo and Xiao, Wendong and Yuan, Liang},
	year = {2024},
}

@inproceedings{sgf,
  title={Scenegraphfusion: Incremental 3d scene graph prediction from rgb-d sequences},
  author={Wu, Shun-Cheng and Wald, Johanna and Tateno, Keisuke and Navab, Nassir and Tombari, Federico},
  booktitle={Proceedings of IEEE/CVF Conference on Computer Vision and Pattern Recognition},
  year={2021}
}

@inproceedings{rosinol_3d_2020,
	title = {{3D} {Dynamic} {Scene} {Graphs}: {Actionable} {Spatial} {Perception} with {Places}, {Objects}, and {Humans}},
	isbn = {978-0-9923747-6-1},
	shorttitle = {{3D} {Dynamic} {Scene} {Graphs}},
	doi = {10.15607/RSS.2020.XVI.079},
	language = {en},
	urldate = {2024-08-01},
	booktitle = {Robotics: {Science} and {Systems} {XVI}},
	author = {Rosinol, Antoni and Gupta, Arjun and Abate, Marcus and Shi, Jingnan and Carlone, Luca},
	year = {2020},
}

@article{behrens_lost_2024,
  author={Behrens, Tjark and Zurbrügg, René and Pollefeys, Marc and Bauer, Zuria and Blum, Hermann},
  journal={IEEE Robotics and Automation Letters}, 
  title={Lost \& Found: Tracking Changes From Egocentric Observations in 3D Dynamic Scene Graphs}, 
  year={2025},
  doi={10.1109/LRA.2025.3544518}}

@inproceedings{wang2016iros,
    AUTHOR     = {John Wang and Edwin Olson},
    TITLE      = {{AprilTag} 2: Efficient and robust fiducial detection},
    BOOKTITLE  = {Proceedings of the {IEEE/RSJ} International Conference on Intelligent
                 Robots and Systems {(IROS)}},
    YEAR       = {2016},
}

@article{bescos2018dynaslam,
  title={DynaSLAM: Tracking, mapping, and inpainting in dynamic scenes},
  author={Bescos, Berta and F{\'a}cil, Jos{\'e} M and Civera, Javier and Neira, Jos{\'e}},
  journal={IEEE Robotics and Automation Letters},
  year={2018},
  publisher={IEEE}
}

@inproceedings{wald2020learning,
  title={Learning 3d semantic scene graphs from 3d indoor reconstructions},
  author={Wald, Johanna and Dhamo, Helisa and Navab, Nassir and Tombari, Federico},
  booktitle={Proceedings of IEEE /CVF Conference on Computer Vision and Pattern Recognition},
  year={2020}
}

@inproceedings{qiu2022airdos,
  title={AirDOS: Dynamic SLAM benefits from articulated objects},
  author={Qiu, Yuheng and Wang, Chen and Wang, Wenshan and Henein, Mina and Scherer, Sebastian},
  booktitle={2022 International Conference on Robotics and Automation (ICRA)},
  year={2022},
  organization={IEEE}
}

@article{song2022dynavins,
  title={DynaVINS: a visual-inertial SLAM for dynamic environments},
  author={Song, Seungwon and Lim, Hyungtae and Lee, Alex Junho and Myung, Hyun},
  journal={IEEE Robotics and Automation Letters},
  year={2022},
  publisher={IEEE}
}

@article{long2021rigidfusion,
  title={Rigidfusion: Robot localisation and mapping in environments with large dynamic rigid objects},
  author={Long, Ran and Rauch, Christian and Zhang, Tianwei and Ivan, Vladimir and Vijayakumar, Sethu},
  journal={IEEE Robotics and Automation Letters},
  year={2021},
  publisher={IEEE}
}

@inproceedings{strecke2019fusion,
  title={Em-fusion: Dynamic object-level slam with probabilistic data association},
  author={Strecke, Michael and Stuckler, Jorg},
  booktitle={Proceedings of the IEEE/CVF International Conference on Computer Vision},
  year={2019}
}

\end{document}